\documentclass[11pt]{article}

\usepackage[final]{acl}

\usepackage{times}
\usepackage{latexsym}
\usepackage[T1]{fontenc}
\usepackage[utf8]{inputenc}
\usepackage{microtype}
\usepackage{inconsolata}

\usepackage{amsmath,amssymb,amsfonts,amsthm}
\usepackage{mathrsfs}
\usepackage{graphicx}
\usepackage{booktabs}
\usepackage{multirow}
\usepackage{makecell}
\usepackage{tabularx}
\usepackage{float}
\usepackage{placeins}
\PassOptionsToPackage{table}{xcolor}
\usepackage{xcolor}
\usepackage{colortbl}  
\usepackage{url}
\usepackage{listings}
\usepackage{tikz}
\usetikzlibrary{arrows.meta,positioning,fit,backgrounds,calc,shapes.geometric,shadings}

\definecolor{lightred}{RGB}{255,200,200}
\definecolor{lightblue}{RGB}{200,220,255}
\definecolor{salmon}{RGB}{250,128,114}
\definecolor{tracemale}{RGB}{14,95,115}     
\definecolor{tracefemale}{RGB}{196,69,54}   

\title{Does Thinking Help Fairness?\\ Reasoning Tokens Resolve Some Biases but Create More}

\author{%
  Deng Pan \\
  University of Notre Dame \\
  \texttt{dpan@nd.edu} \And
  Joe Germino \\
  University of Notre Dame \\
  \texttt{jgermino@nd.edu} \And
  Yihong Ma \\
  University of Notre Dame \\
  \texttt{yihongma97@gmail.com} \AND
  Elizabeth Daly \\
  IBM Research, Dublin \\
  \texttt{elizabeth.daly@ie.ibm.com} \And
  Nuno Moniz \\
  University of Notre Dame \\
  \texttt{nunomoniz@nd.edu} \And
  Ting Hua \\
  University of Notre Dame \\
  \texttt{thua@nd.edu} \AND
  Nitesh Chawla \\
  University of Notre Dame \\
  \texttt{nchawla@nd.edu}
}

\begin{document}

\maketitle

\begin{abstract}
Thinking in reasoning language models (RLMs) has been subject to debate on whether it resolves or amplifies bias. Prior works have shown competing conclusions in both directions.
Using a within-model thinking-vs.-non-thinking ablation across QwQ-32B, DeepSeek-R1-Distill-Qwen-32B, and Qwen3-32B on three high-stakes decision tasks (Adult, COMPAS, Credit), we show that thinking has an asymmetric dual effect on counterfactual fairness: it both resolves counterfactual flips produced by the non-thinking baseline and creates new flips at near-saturating model confidence. In all nine (model, dataset) combinations, the created flips outnumber the resolved flips by roughly $5\times$.
To explain the effect, we treat the thinking trace itself as a measurable site of fairness change and study it through two dynamic instruments: 1) We propose Counterfactual Depth Probability Gap (CDPG) to track bias evolution along thinking depth, and observe that bias propagates and amplifies with thinking. 2) We also formulate the Bias Transition Matrix (BTM) to show how predictions of counterfactual pairs change from non-thinking to thinking, and find that the asymmetric dual effect originates in the pair-state joint transition. 
\footnote{Code, prompts, and all model outputs are released at \url{https://github.com/pd90506/fairness_audit}.}
\end{abstract}

\section{Introduction}\label{sec:intro}

Reasoning language models (RLMs) are increasingly deployed in high-stakes decision pipelines, in part because their visible \emph{thinking traces}, enclosed in \texttt{<think>}\ldots\texttt{</think>} tags, are advertised as making deliberation transparent and auditable~\citep{guo2025deepseek,yang2025qwen3,qwq32b,zhao2024explainability}. These deployments raise immediate fairness concerns: LLM decisions reproduce historical disparities along protected attributes, and any deliberative step that intervenes between input and decision is itself a locus of fairness risk~\citep{tamkin2023evaluating,gallegos2024bias,caton2024fairness}. Whether the thinking phase mitigates these disparities or amplifies them is a critical question that needs to be addressed.

Prior work reaches opposing conclusions on this question, ultimately because the studies in conflict measure different fairness concepts on different tasks. One line reports that explicit thinking reduces bias on aggregate stereotype probes~\citep{kabra2025regift}; a related implicit-association reading~\citep{apsel2026inferencebias} measures bias via Implicit Association Test (IAT) style instruments and similarly argues for a reducing effect. 
A second line reports that thinking amplifies bias on stereotype QA benchmarks~\citep{ADBP2025}.
There is also a parallel line of study on adversarial injection, which shows that adversarial prompts can plant bias inside the trace~\citep{cantini2025reasoningbias,turpin2023language}.

\begin{figure*}[ht]
    \centering
    \resizebox{\textwidth}{!}{
\begin{tikzpicture}[
    font=\small,
    box/.style={
        draw, rounded corners=4pt, inner sep=6pt, align=center, line width=0.5pt
    },
    qbox/.style={box, minimum width=32mm, minimum height=20mm},
    tracebox/.style={
        draw, rounded corners=5pt, inner sep=6pt, align=left,
        text width=60mm, font=\small, line width=0.4pt
    },
    ansbox/.style={box, minimum width=22mm, minimum height=14mm},
    arrow/.style={-{Stealth[length=2.5mm]}, thick, gray!50},
    dasharrow/.style={-{Stealth[length=2.2mm]}, dashed, gray!40, thin},
    secttitle/.style={font=\normalsize\scshape\bfseries, gray!40!black},
]

\node[secttitle] at (6.5,3.0) {Thinking Arm};

\node[qbox, fill=blue!8, draw=blue!40!black] (qa) at (0,1.7) {%
    $Q^a$\\[3pt]
    {\scriptsize age 56, tech support,}\\[-1pt]
    {\scriptsize local-gov, bachelor's}\\[4pt]
    {\normalsize\bfseries sex = Male}
};

\node[tracebox, fill=yellow!8, draw=yellow!40!black] (ta) at (6.5,1.7) {%
    ``Putting it all together: {\bfseries\color{blue!55!black} it's possible his salary is around \$50K or slightly above.} So I might lean towards `Yes'.''
};

\node[ansbox, fill=blue!8, draw=blue!40!black] (pa) at (12.5,1.7) {%
    {\Large\color{blue!60!black}\bfseries Yes}\\[3pt]
    {\small $P^a{=}0.83$}
};

\node[qbox, fill=red!7, draw=red!35!black] (qb) at (0,-0.5) {%
    $Q^{a'}$\\[3pt]
    {\scriptsize age 56, tech support,}\\[-1pt]
    {\scriptsize local-gov, bachelor's}\\[4pt]
    {\normalsize\bfseries sex = Female}
};

\node[tracebox, fill=yellow!8, draw=yellow!40!black] (tb) at (6.5,-0.5) {%
    ``Wait, but she's in Local-gov. Public sector salaries are often lower than private. {\bfseries\color{red!55!black} So even with a Bachelors, her salary might be lower.} \ldots\ So maybe the answer is No.''
};

\node[ansbox, fill=red!8, draw=red!40!black] (pb) at (12.5,-0.5) {%
    {\Large\color{red!60!black}\bfseries No}\\[3pt]
    {\small $P^{a'}{=}0.01$}
};

\node[box, fill=violet!10, draw=violet!45!black, minimum width=26mm, minimum height=18mm]
    (cdpg) at (16,0.6) {%
    $\mathrm{CDPG}_K$\\[4pt]
    {\Large\bfseries\color{violet!70!black} $= 0.82$}
};

\draw[arrow] (qa) -- (ta);
\draw[arrow] (qb) -- (tb);
\draw[arrow] (ta) -- (pa);
\draw[arrow] (tb) -- (pb);
\draw[dasharrow] (pa.east) -- ([yshift=4mm]cdpg.west);
\draw[dasharrow] (pb.east) -- ([yshift=-4mm]cdpg.west);

\draw[gray!25, line width=0.5pt] (-2.2,-2.0) -- (18,-2.0);

\node[secttitle] at (6.5,-2.45) {Non-thinking Arm};

\node[box, fill=blue!8, draw=blue!35!black, minimum width=16mm, minimum height=8mm]
    (nqa) at (2,-3.1) {\small $Q^a$};
\node[box, fill=green!12, draw=green!45!black, minimum width=18mm, minimum height=8mm]
    (nya) at (4.8,-3.1) {\normalsize\bfseries Yes};

\node[box, fill=red!7, draw=red!35!black, minimum width=16mm, minimum height=8mm]
    (nqb) at (8.5,-3.1) {\small $Q^{a'}$};
\node[box, fill=green!12, draw=green!45!black, minimum width=18mm, minimum height=8mm]
    (nyb) at (11.3,-3.1) {\normalsize\bfseries Yes};

\draw[arrow, thin] (nqa) -- (nya);
\draw[arrow, thin] (nqb) -- (nyb);

\end{tikzpicture}}
    \caption{\textbf{Anatomy of a cell-$c$ pair (bias created by thinking).} Qwen3-32B on Adult. The two sides agree under the non-thinking arm but diverge under thinking ($\mathrm{CDPG}_K{=}0.82$); highlighted spans are the bias-relevant phrasing.}
    \label{fig:contingency_flow}
\end{figure*}
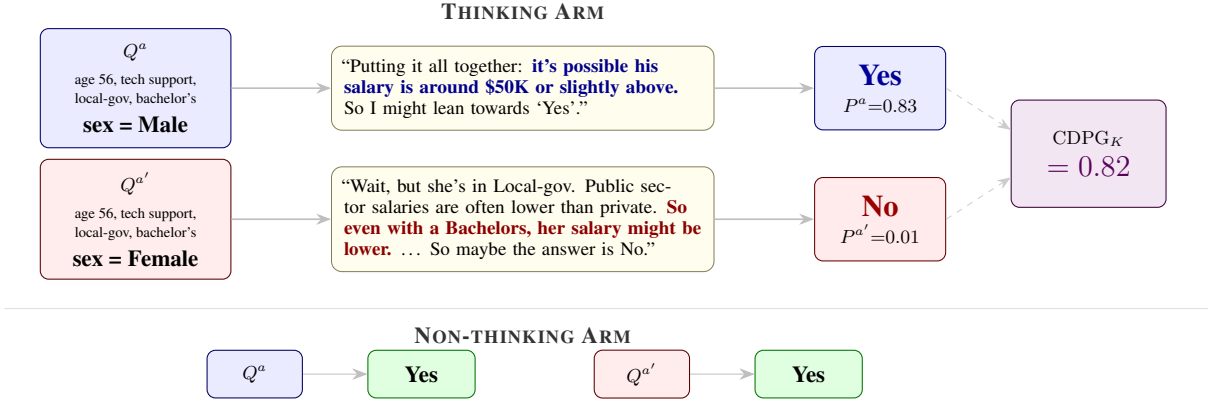

Mixed conclusions on the same question leave practitioners without a defensible standard for deploying RLMs in fairness-sensitive settings. We argue that the disagreement runs deeper than metric choice: all prior audits evaluate only the final prediction, leaving the thinking process itself unexamined. Instead, we study the \emph{dynamics} of \textbf{how bias propagates} during thinking.

Across three open-weight $32$B-class RLMs and three tabular decision tasks, an outcome-level audit reveals an \emph{asymmetric dual effect} in all 9/9 (model, dataset) scenarios: thinking creates roughly $5\times$ more new counterfactual bias pairs than it resolves. Figure~\ref{fig:contingency_flow} sketches a concrete instance: a counterfactual pair that agrees without thinking but diverges under thinking, with the trace revealing gender-biased reasoning.

To explain this asymmetry, we propose two complementary dynamic instruments: Counterfactual Depth Probability Gap (CDPG) to audit bias propagation during thinking, and Bias Transition Matrix (BTM) to track instance-level bias evolution from non-thinking to thinking. CDPG reveals that bias propagates and amplifies with thinking depth; BTM reveals that the asymmetry originates in the pair-state joint transition.

\paragraph{Contributions.}
\begin{enumerate}\itemsep0pt
    \item We \textbf{identify an asymmetric dual effect} of thinking on counterfactual fairness: across all nine (model, dataset) settings, thinking simultaneously resolves a subset of counterfactual flips produced by non-thinking inference and creates roughly $5\times$ more new flips at high confidence. (\S\ref{subsec:phenomenon})
    \item We \textbf{propose two dynamic instruments}, Counterfactual Depth Probability Gap (CDPG) and Bias Transition Matrix (BTM), to surface structure invisible to outcome-only audits. (\S\ref{sec:methods})
    \item We \textbf{reveal an anchor/mover dynamic} within the thinking trace: one side of a counterfactual pair maintains its initial prediction while the other drifts toward the opposite boundary, explaining how thinking sharpens both fair and biased outcomes in lockstep. (\S\ref{subsec:propagation})
    \item We \textbf{trace the structural origin of the dual effect} to the pair-state joint transition: the group-level transition is nearly identical across protected sides, but the pair-state transition is asymmetric, and that asymmetry drives creation to outweigh resolution. (\S\ref{subsec:transition})
\end{enumerate}

\section{Related Work}\label{sec:related-work}

\subsection{Fairness audits of LLMs}\label{subsec:fairness-audits}
Empirical fairness audits of LLMs descend from two distinct traditions. Group-level audits operationalise demographic-parity and equalised-odds criteria from algorithmic fairness \citep{hardt_equality_2016}, consolidated for LLMs in recent surveys \citep{gallegos2024bias}. Counterfactual audits compare an instance's outcome under different values of a protected attribute \citep{garg2019counterfactual,tamkin2023evaluating}. We take counterfactual disagreement as the primary signal and report group fairness alongside it to expose where the two diagnostics agree and where they diverge.

\subsection{Competing reports on reasoning and fairness}\label{subsec:competing-reports}
On-topic prior work falls into three lines: two camps that disagree on the directional effect of thinking on fairness (mitigation vs.\ amplification), and a separate line on deliberate bias injection.

\paragraph{Thinking as bias mitigation.} This line reports that explicit thinking reduces bias. \citet{kabra2025regift} report that reasoning-trained models exhibit lower stereotypical bias on standard benchmarks, reading the change as a net reduction. The implicit-association reading of \citet{apsel2026inferencebias} is adjacent but conceptually distinct: it measures implicit bias in the IAT sense rather than explicit decision-level bias. \citet{cherepanova2025improving} use CoT self-refinement as one of four interventions to improve demographic parity on tabular fairness datasets, illustrating that CoT can be used as a deliberate mitigation lever.

\paragraph{Thinking as bias amplification.} This line reports that thinking amplifies bias. \citet{shaikh2023second} show that zero-shot chain-of-thought prompting elevates harmful and stereotypical generation on socially sensitive questions relative to a no-CoT baseline. \citet{ADBP2025} report the same direction on BBQ stereotype QA~\citep{parrish2022bbq} for reasoning models. \citet{karvonen2025robustly} run a counterfactual hiring audit on a reasoning-trained model and find 10--12\% interview-rate disparities across race and gender, with the disparity hidden inside the CoT trace. \citet{moore2024reasoning} extend counterfactual prompting to CoT on general-purpose LLMs (not reasoning models) and find that CoT alone fails to mitigate, and can reinforce, base-rate-driven cross-group bias.

\paragraph{Adversarial paradigm: bias injection (out of scope).} A separate line probes how reasoning models behave once bias has been deliberately implanted via jailbreak prompts or injected biasing features \citep{cantini2025reasoningbias,turpin2023language}; we treat it as out of scope since it characterises implanted rather than emergent bias.

The opposing conclusions of the mitigation and amplification camps motivate a finer-grained probing of the thinking process itself, to understand the mechanism by which thinking can both mitigate bias in some cases and amplify it in others.

\subsection{Probing reasoning trace dynamics}\label{subsec:probing-traces}
Probing methods treat the trace itself as a measurable object. \citet{lanham2023faithfulness} introduce prefix-truncation probing as a faithfulness test for chain-of-thought (``early answering''): the trace is truncated at varying lengths and the model is re-prompted for an answer at each truncation, testing whether the stated reasoning is causally load-bearing on the final answer. \citet{ADBP2025} apply the same prefix-truncation construction to stereotype QA on BBQ: they re-complete truncated reasoning prefixes and detect within-trace answer shifts as a bias proxy, then use those shifts as a re-voting signal. Our bias-propagation instrument (\S\ref{subsec:cdpg-tool}) builds on this prefix-truncation construction, and our pair-state transition matrix (\S\ref{subsec:transition-tools}) complements it by characterising the nothink-to-think joint dynamics at the pair level.

\section{Methods}\label{sec:methods}

\subsection{Fairness Definitions}\label{subsec:fairness-defs}

We take counterfactual disagreement as the primary fairness signal. A counterfactual pair $(x_i(A=\alpha), x_i(A=\alpha'))$ consists of two instances that differ only in the protected attribute $A$. We quantify disagreement on each pair in two ways. The hard \textbf{counterfactual flip rate}
\begin{align}
\mathcal{D}_{\text{cf}} &= \tfrac{1}{N}\sum_{i} \mathbf{1}\bigl[\hat{y}_i(\alpha) \neq \hat{y}_i(\alpha')\bigr]
\label{eq:dcf}
\end{align}
counts the fraction of pairs whose binary predictions disagree. The soft \textbf{counterfactual disparity}
\begin{align}
\mathcal{D}_{\text{disp}}^{(i)} &= \bigl|P(\hat{y}{=}1\mid x_i(\alpha)) - P(\hat{y}{=}1\mid x_i(\alpha'))\bigr|
\label{eq:ddisp}
\end{align}
is the per-pair probability gap; we read it at the end of the thinking trace through $|\text{CDPG}_K^{(i)}|$ (\S\ref{subsec:cdpg-tool}). We follow \citet{garg2019counterfactual} and \citet{tamkin2023evaluating} in operationalising counterfactual fairness as prompt-level substitution of the protected attribute. 

The \textbf{group disparity}~\citep{hardt_equality_2016,fairlearn,gallegos2024bias} is the demographic-parity difference between two protected-attribute populations,
\begin{align}
\mathcal{D}_{\text{group}} &= \bigl|P(\hat{y}{=}1\mid g_i) - P(\hat{y}{=}1\mid g_j)\bigr|.
\label{eq:dgroup}
\end{align}
Counterfactual fairness and group fairness capture different harms and can conflict.

\subsection{Within-pair Contingency Cells}\label{subsec:contingency-cells}

The contingency cells defined here are our primary instrument for surfacing the asymmetric dual effect of thinking on counterfactual fairness (\S\ref{subsec:phenomenon}). For each counterfactual pair we cross-tabulate the flip indicator under thinking against the flip indicator under the within-model non-thinking arm. The resulting four cells are labelled $\{a,b,c,d\}$ (Table~\ref{tab:cell_definitions}): cell $a$ contains pairs with no flip in either arm; cell $b$ contains pairs flipped under non-thinking but agreeing under thinking, the \emph{flips resolved by thinking}; cell $c$ contains pairs flipped under thinking but agreeing under non-thinking, the \emph{flips created by thinking}; cell $d$ contains pairs flipped under both arms. Figure~\ref{fig:contingency_flow} illustrates a concrete cell-$c$ instance: a counterfactual pair that agrees under the non-thinking arm but diverges under thinking, with the trace itself surfacing biased reasoning.

\begin{table}[h]
\centering
\renewcommand{\arraystretch}{1.3}
\setlength{\tabcolsep}{6pt}
\small
\begin{tabular}{@{}c c | c c@{}}
\toprule
 & & \multicolumn{2}{c}{\textbf{Think}} \\
 \cmidrule(lr){3-4}
 & & \emph{no flip} & \emph{flip} \\
\midrule
\multirow{2}{*}{\textbf{No-think}}
 & \emph{no flip} &
\cellcolor{yellow!14}$a$: both agree &
\cellcolor{salmon!22}$c$: created \\
 & \emph{flip} &
\cellcolor{teal!20}$b$: resolved &
\cellcolor{purple!22}$d$: both flip \\
\bottomrule
\end{tabular}
\caption{\textbf{Within-pair contingency cells.} Each counterfactual pair is classified by whether a flip ($\hat{y}^a \neq \hat{y}^{a'}$) occurs under the thinking and non-thinking arms. 
}
\label{tab:cell_definitions}
\end{table}

\subsection{Bias Propagation Tool: \texorpdfstring{$\text{CDPG}_k$}{CDPG\_k}}\label{subsec:cdpg-tool}

The Counterfactual Depth Probability Gap at depth $k$, $\text{CDPG}_k$, is our instrument for observing how counterfactual bias evolves along thinking depth: a per-pair signal that quantifies the counterfactual decision gap after the first $k$ segments of the thinking trace. The construction adapts the prefix-truncation probing scaffold of \citet{lanham2023faithfulness}, originally applied as an early-answering faithfulness test on chain-of-thought, to counterfactually paired inputs.

For a counterfactual query pair $(Q^\alpha,Q^{\alpha'})$ with thinking traces $T^\alpha$ and $T^{\alpha'}$, we partition each trace into $K{=}10$ equal-token segments, $T^\alpha = (S^\alpha_1,\ldots,S^\alpha_K)$ and likewise for $T^{\alpha'}$. For each depth $k\in\{1,\ldots,K\}$ and attribute value $A\in\{\alpha,\alpha'\}$ we re-feed the prefix $(Q^A,S^A_{1:k},\text{suffix})$ to the model, where the suffix is the answer-elicitation template that closes the thinking block and opens the answer position (Appendix~\ref{app:prompt_format}). At the answer position we extract the logits for the \texttt{<Yes>} and \texttt{<No>} tokens only and apply softmax over this pair (not over the full vocabulary), giving
\begin{align}
P^\alpha_k &= P(\hat{y}{=}1\mid Q^\alpha,S^\alpha_{1:k}), \\
P^{\alpha'}_k &= P(\hat{y}{=}1\mid Q^{\alpha'},S^{\alpha'}_{1:k}).
\label{eq:pk}
\end{align}
and the per-pair, per-depth signal
\begin{align}
\text{CDPG}_k^{(i)} &= P^\alpha_k - P^{\alpha'}_k.
\label{eq:cdpg}
\end{align}
At $k{=}K$, $|\text{CDPG}_K^{(i)}|$ equals the thinking-arm per-pair counterfactual disparity $\mathcal{D}_{\text{disp}}^{(i)}$, so the $k{=}1{:}K$ progression decomposes the final disparity over reasoning depth. Figure~\ref{fig:pipeline} illustrates the full pipeline.

\paragraph{Interventional reading.}
Re-feeding the truncated prefix as a fresh context is what makes $P^A_k$ interventional: it answers what decision the model would commit to if thinking were interrupted at depth $k$ and the model were forced to answer immediately.

\paragraph{Aggregation.}
For pooled or stratified reporting we average $|\text{CDPG}_k^{(i)}|$ over pairs within a chosen subpopulation $\mathcal{S}$ (a contingency cell, a (model, dataset) cell, or the full pair pool), giving
$\overline{|\text{CDPG}_k|} = \tfrac{1}{|\mathcal{S}|}\sum_{i\in\mathcal{S}} |\text{CDPG}_k^{(i)}|$.
This aggregation is used to build the per-cell trajectories in \S\ref{subsec:propagation}.

\subsection{Transition Matrix Tools}\label{subsec:transition-tools}

Our second dynamic instrument, the \textbf{Bias Transition Matrix (BTM)}, tracks how predictions move between the non-thinking and thinking arms at two granularities: a group-level matrix $T^A$ that conditions on the protected attribute alone, and a pair-state matrix $T^{\text{pair}}$ that tracks the joint state of a counterfactual pair.

\paragraph{Group-level transition matrix $T^A$.}
Each instance yields a non-thinking prediction $\hat y_{\text{nothink}} \in \{Y,N\}$ and a thinking prediction $\hat y_{\text{think}} \in \{Y,N\}$, where $Y,N$ represents Yes and No, respectively. Conditioning on the protected attribute $A$, the group-level transition matrix is
\begin{align}
T^A_{ij} &= P(\hat y_{\text{think}}{=}j \mid \hat y_{\text{nothink}}{=}i,\, A),
\label{eq:tinst}
\end{align}
for $i,j\in\{Y,N\}$ and $A\in\{\alpha,\alpha'\}$, estimated by empirical conditional frequency. $T^A$ characterises how individual predictions move from non-thinking to thinking for each protected attribute value, without reference to the counterfactual partner.

\paragraph{Pair-state transition matrix $T^{\text{pair}}$.}
Each counterfactual pair yields a non-thinking pair-state $s_{\text{nt}} \in \{\text{NN},\text{YY},\text{M}\}$ and a thinking pair-state $s_{\text{th}} \in \{\text{NN},\text{YY},\text{M}\}$, where NN denotes both sides predicted negative, YY both sides predicted positive, and M a mixed state (equivalently a counterfactual flip). The pair-state transition matrix is
\begin{align}
T^{\text{pair}}_{ij} &= P(s_{\text{th}}{=}j \mid s_{\text{nt}}{=}i),
\label{eq:tpair}
\end{align}
for $i,j\in\{\text{NN},\text{YY},\text{M}\}$, estimated by empirical conditional frequency over the pooled pair population, and also stratified per (model, dataset) cell. $T^{\text{pair}}$ characterises how the joint pair distribution moves between the two arms at the pair level. The contingency cells of \S\ref{subsec:contingency-cells} are off-diagonal blocks of $T^{\text{pair}}$: cell $c$ (created flips) is the $\{\text{NN},\text{YY}\}{\to}\text{M}$ mass, cell $b$ (resolved flips) is the $\text{M}{\to}\{\text{NN},\text{YY}\}$ mass, and the resolve-vs-create asymmetry of the dual effect is the asymmetry between these two off-diagonal blocks. By comparison with the group-level $T^A$ above, $T^{\text{pair}}$ captures the joint dynamics that $T^A$ marginalises out: if $T^\alpha$ and $T^{\alpha'}$ are nearly identical, any asymmetry in the dual effect must originate at the joint pair-state level captured by $T^{\text{pair}}$.

\section{Results}\label{sec:results}

\subsection{Experimental setup}\label{subsec:setup}
We evaluate three high-stakes tabular decision datasets. \textbf{Adult}~\citep{adult_2} predicts whether an individual's annual income exceeds \$50K, with \emph{sex} as the protected attribute. \textbf{COMPAS}~\citep{angwin2022machine} predicts two-year recidivism, with \emph{race} as the protected attribute (binarised to White vs.\ Non-White). \textbf{Credit}~\citep{default_of_credit_card_clients_350} predicts next-month credit-card default, with \emph{sex} as the protected attribute. To prevent confounding between protected attributes, we drop the race column from Adult and the sex column from COMPAS so that each dataset isolates a single protected attribute under counterfactual substitution; Credit contains no second protected column. Each dataset is sampled at $N{=}5{,}000$ records and each record is paired with its protected-attribute counterfactual via fixed-template natural-language profiling (Appendix~\ref{app:nl_profiles}). 

We evaluate three open-weight $32$B-class reasoning models that expose thinking tokens natively: QwQ-32B~\citep{qwq32b}, DeepSeek-R1-Distill-Qwen-32B~\citep{guo2025deepseek}, and Qwen3-32B~\citep{yang2025qwen3}. The within-model non-thinking baseline is implemented by inserting a closed empty \texttt{<think></think>} block. All models run at 4-bit AWQ quantisation on A100 80\,GB GPUs with greedy decoding ($T{=}0$); CDPG force-answer probabilities are read directly from answer-token log-probabilities.

We retain only counterfactual pairs whose Yes/No answer parses on both sides of both arms; the effective $N$ falls below $5{,}000$ when the model emits no Yes/No token and therefore cannot be classified. The most severe drop is DeepSeek-R1-Distill-Qwen-32B on COMPAS ($N{=}1{,}430$), where the non-thinking arm refuses to answer on the majority of records; the remaining cases lie within a few records of $5{,}000$ (Table~\ref{tab:dual_effect_main}).

\subsection{The phenomenon: asymmetric dual effect of thinking}\label{subsec:phenomenon}

\paragraph{Group disparity gives an inconsistent signal.}
\begin{table*}[ht]
\centering
\renewcommand{\arraystretch}{1.15}
\setlength{\tabcolsep}{3.5pt}
\small
\begin{tabular}{l l r r r r r r r r r r}
\toprule
& & & \multicolumn{4}{c}{\textbf{Contingency cells (counts)}} & \textbf{McNemar} & \multicolumn{2}{c}{$\overline{\mathcal{D}_{\text{disp}}}$ \textbf{in cell $b$}} & \multicolumn{2}{c}{$\overline{\mathcal{D}_{\text{disp}}}$ \textbf{in cell $c$}} \\
\cmidrule(lr){4-7}\cmidrule(lr){8-8}\cmidrule(lr){9-10}\cmidrule(lr){11-12}
\textbf{Model} & \textbf{Dataset} & $N$ & $a$ & $b$ & $c$ & $d$ & $p$ & nothink & think$_K$ & nothink & think$_K$ \\
\midrule
\multirow{3}{*}{Qwen3-32B} & Adult  & 5000 & 4531 &  75 & 362 & 32 & $3.7{\times}10^{-46}$ & 0.219 & 0.048 & 0.055 & \textbf{0.859} \\
                          & COMPAS & 5000 & 4511 & 143 & 344 &  2 & $3.7{\times}10^{-20}$ & 0.150 & 0.037 & 0.046 & \textbf{0.781} \\
                          & Credit & 4996 & 4867 &  23 & 105 &  1 & $1.0{\times}10^{-13}$ & 0.075 & 0.012 & 0.023 & \textbf{0.755} \\
\midrule
\multirow{3}{*}{QwQ-32B}  & Adult  & 5000 & 4396 & 152 & 443 &  9 & $6.1{\times}10^{-34}$ & 0.131 & 0.001 & 0.029 & \textbf{0.980} \\
                          & COMPAS & 5000 & 4634 &  15 & 351 &  0 & $2.3{\times}10^{-84}$ & 0.262 & 0.008 & 0.002 & \textbf{0.891} \\
                          & Credit & 4984 & 4265 & 121 & 595 &  3 & $5.3{\times}10^{-76}$ & 0.156 & 0.016 & 0.007 & \textbf{0.642} \\
\midrule
\multirow{3}{*}{R1-Distill-Qwen-32B}
                          & Adult  & 4999 & 4255 &  77 & 646 & 21 & $7.6{\times}10^{-113}$ & 0.076 & 0.026 & 0.032 & \textbf{0.918} \\
                          & COMPAS & 1430 & 1247 &   4 & 179 &  0 & $7.5{\times}10^{-48}$ & 0.085 & 0.003 & 0.042 & \textbf{0.933} \\
                          & Credit & 4995 & 4502 &  10 & 483 &  0 & $1.7{\times}10^{-128}$ & 0.034 & 0.000 & 0.003 & \textbf{0.791} \\
\midrule
\textbf{Pooled} & \textbf{(9 cells)} & 41404 & 37208 & 620 & 3508 & 68 & $<10^{-485}$ & 0.143 & 0.022 & 0.024 & \textbf{0.835} \\
\bottomrule
\end{tabular}
\caption{\textbf{Dual-effect contingency table with within-pair gap reshaping in cells $b$ and $c$.} Cell counts $a,b,c,d$; the McNemar column reports the exact two-sided $p$-value of McNemar's test on the discordant cells $(b,c)$, rejecting $H_0{:}\,b{=}c$ in every setting; rightmost columns give the mean per-pair counterfactual disparity $\overline{\mathcal{D}_{\text{disp}}}$ in cells $b,c$ under the nothink and final-thinking arms ($\mathcal{D}_{\text{disp}}^{(i)} \equiv |\mathrm{CDPG}_K^{(i)}|$, Eqs.~\ref{eq:ddisp},~\ref{eq:cdpg}). $N$ retains pairs parsing on both sides of both arms.}
\label{tab:dual_effect_main}
\end{table*}

\begin{figure*}[t]
    \centering
    \includegraphics[width=\textwidth]{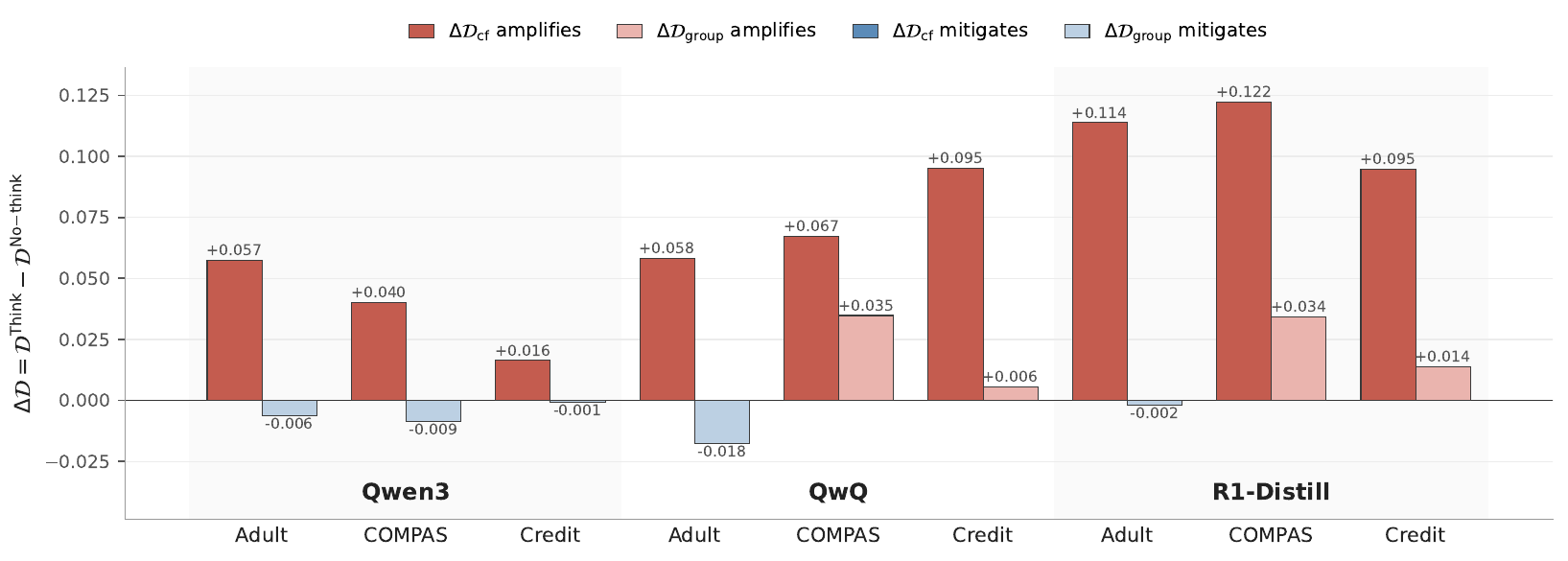}
    \caption{\textbf{Per-cell change under thinking of counterfactual disparity ($\Delta\mathcal{D}_{\mathrm{cf}}$, Eq.~\ref{eq:dcf}) vs.\ group disparity ($\Delta\mathcal{D}_{\mathrm{group}}$, Eq.~\ref{eq:dgroup}).} Absolute Think/No-think values are reported in Table~\ref{tab:d_group} (Appendix~\ref{app:disparity_values}).}
    \label{fig:delta_paired_bars}
\end{figure*}
Figure~\ref{fig:delta_paired_bars} shows that $\Delta\mathcal{D}_{\text{group}}$ (pale bars) has no consistent sign across the nine (model, dataset) scenarios: thinking amplifies group disparity in four cases, mitigates it in four, and leaves it effectively unchanged in one. The magnitudes are also small ($|\Delta\mathcal{D}_{\text{group}}| \leq 0.035$, with six cases at $\leq 0.015$). On QwQ-Adult $\Delta\mathcal{D}_{\text{group}} = -0.018$ (a single-group audit reads ``thinking helps''); on QwQ-COMPAS $\Delta\mathcal{D}_{\text{group}} = +0.035$ (the same audit reads ``thinking hurts''). The direction of the group-disparity verdict is dataset-dependent.
This observation, complementing with Section~\ref{subsec:transition}, explains why prior work reaches opposing conclusions in fairness implications. 

\paragraph{Counterfactual disparity is consistently amplified.}
By contrast, $\Delta\mathcal{D}_{\text{cf}}$ (dark bars in Figure~\ref{fig:delta_paired_bars}) is positive in all $9/9$ cells, ranging from $+0.016$ (Qwen3-Credit) to $+0.122$ (R1-Distill-COMPAS). Counterfactual flip rate increases under thinking in every setting tested. To understand the source of this consistent amplification, a finer-grained diagnosis is needed.

\paragraph{Two direct contributors to the counterfactual amplification.}
Table~\ref{tab:dual_effect_main} decomposes the counterfactual amplification into two contributors. \emph{(i) Counterfactual bias creation outnumbers resolution}: $|c| > |b|$ in all nine scenarios, with the pooled gap $3508$ vs $620$ ($\approx 5.7\times$). The asymmetry is statistically decisive: $b$ and $c$ are the discordant cells of the paired thinking-vs-nothink contingency, and McNemar's exact test rejects $H_0{:}\,b{=}c$ in every setting ($p \leq 10^{-13}$ per cell; pooled $p < 10^{-485}$; Table~\ref{tab:dual_effect_main}). Since decoding is greedy ($T{=}0$), each flip is a deterministic function of the trace, so the excess of created over resolved flips cannot be attributed to sampling variance. Nor is the effect specific to the Qwen lineage shared by the three main models: it replicates on gemma-2-27B and Llama-3.1-8B with prompt-elicited traces, at comparable ratios (Appendix~\ref{app:generalization}). \emph{(ii) Within cell $c$, the per-pair counterfactual disparity is strengthened}: the mean $\overline{\mathcal{D}_{\text{disp}}}$ (Eq.~\ref{eq:ddisp}) on cell-$c$ pairs rises from $0.024$ under nothink to $0.835$ under thinking ($\approx 34\times$ pooled). Thinking thus creates more flips and commits to each created flip with markedly higher confidence; the cell-$b$ resolution ($\overline{\mathcal{D}_{\text{disp}}}$ $0.143 \to 0.022$, $6.5\times$) is real but per-pair $\approx 5\times$ weaker than cell-$c$ creation.

\paragraph{What remains for the dynamic analysis.}
Two questions follow from the dual-effect phenomenon and motivate the rest of this section. (i) Where in the thinking trace, and via what mechanism, does the dual effect emerge? (ii) Where, mechanistically, does the asymmetric transition arise that amplifies the counterfactual bias?

\subsection{Bias propagation: the source of the dual effect}\label{subsec:propagation}

\begin{figure*}[ht]
    \centering
    \includegraphics[width=\textwidth]{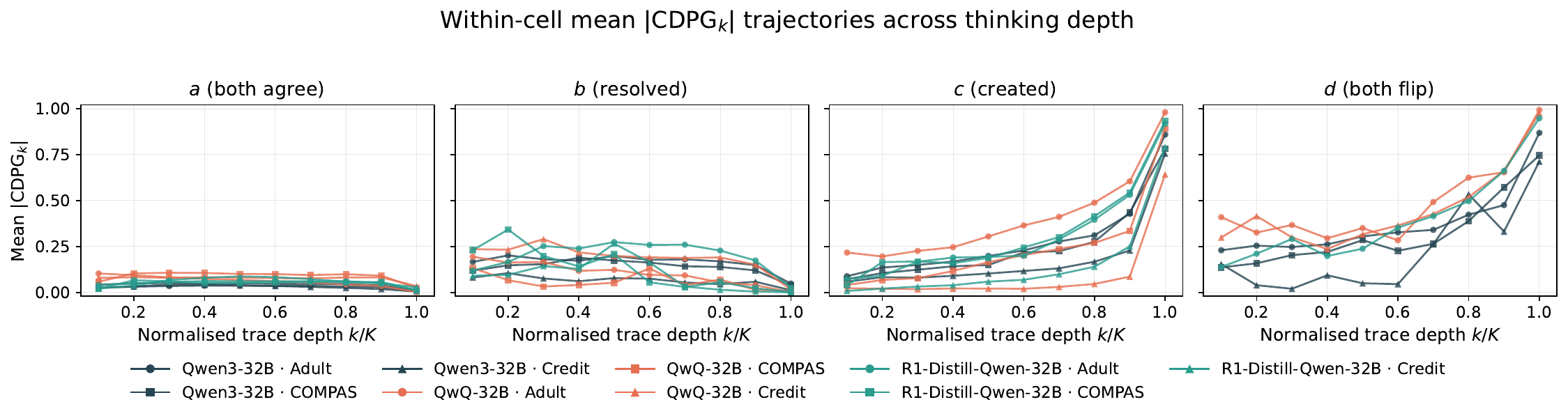}
    \caption{\textbf{The dual effect plays out consistently as the thinking trace deepens.} Mean $|\mathrm{CDPG}_k|$ as a function of normalised thinking depth $k/K$, computed within each contingency cell, pooled across the nine (model, dataset) settings.}
    \label{fig:copg_trajectories}
\end{figure*}

\paragraph{Bias dynamics.}
Figure~\ref{fig:copg_trajectories} plots the per-cell mean $|\text{CDPG}_k|$ against normalised depth $k/K$ for each of the nine (model, dataset) settings. The four cells split along a single axis: whether or not biased reasoning surfaces during thinking. Cells $c$ and $d$, the two cells in which thinking ends with a counterfactual flip, see $|\text{CDPG}_k|$ rise toward saturation: once a bias signal surfaces mid-trace, the remaining depth amplifies it into a high-confidence disagreement. Cells $a$ and $b$, the two cells in which thinking ends with agreement, see $|\text{CDPG}_k|$ stay near zero or decay toward it: when no biased reasoning surfaces, the same amplifier sharpens the prediction toward a confident fair outcome. The four trajectories are thus governed by one mechanism with two visible faces: \textit{thinking commits each pair-side ever more confidently toward whichever boundary it is already drifting to}, and the contingency cell only records whether the two sides land on the same or opposite boundaries. These trajectories are robust to the segmentation granularity: recomputing at $K \in \{5, 10, 20\}$ preserves the cell-$c$ propagation slope's sign and order of magnitude (Appendix~\ref{app:k-sensitivity}).

\begin{figure*}[!t]
    \centering
    \includegraphics[width=0.8\textwidth]{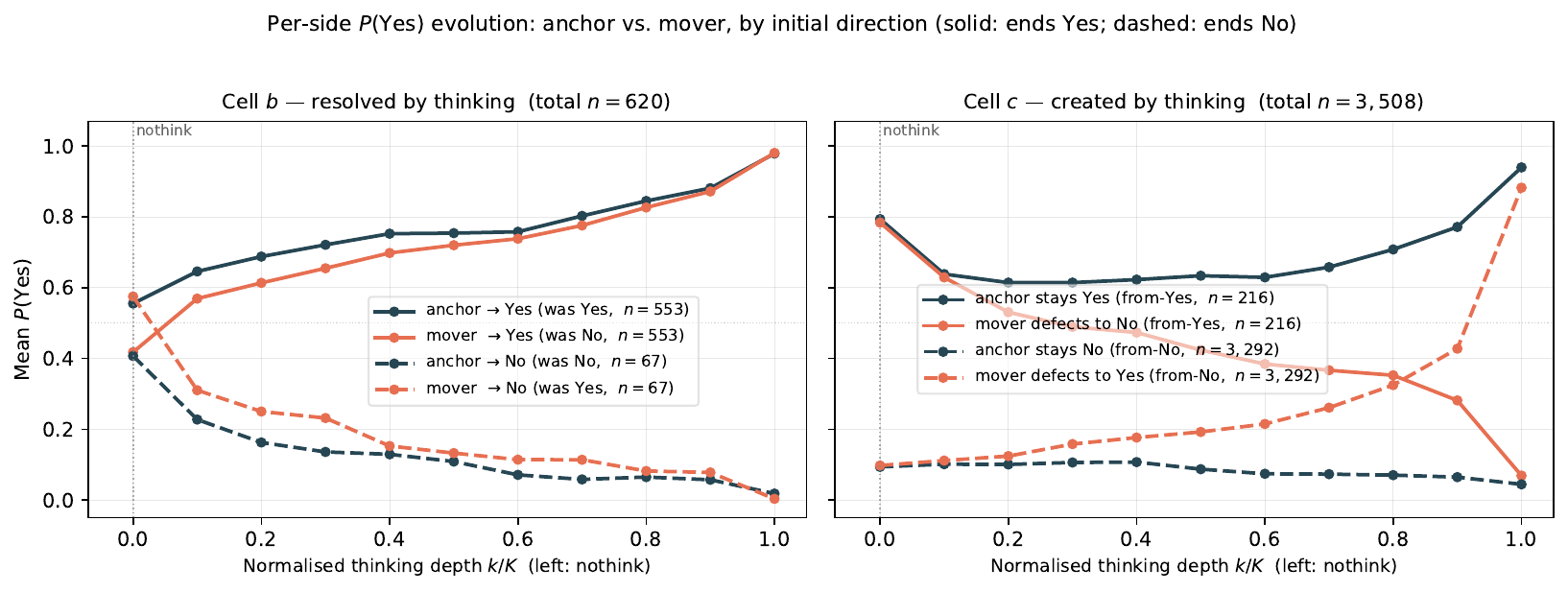}
    \caption{\textbf{Per-side $P(\mathrm{Yes})$ evolution through the thinking trace.} Cell $b$ (left, resolved by thinking) and cell $c$ (right, created by thinking). The leftmost point ($k{=}0$) is the nothink baseline. Pooled across the nine (model, dataset) settings.}
    \label{fig:pyes_evolution}
\end{figure*}

\paragraph{Confidence dynamics.}
To understand where the bias dynamics in $|\text{CDPG}_k|$ come from, we decompose them into the two per-side $P(\text{Yes})$ trajectories that produce them (Figure~\ref{fig:pyes_evolution}, cells $b$ and $c$). Within each cell, one side maintains its initial prediction throughout the trace (the \emph{anchor}) and the other changes prediction (the \emph{mover}). In cell $b$ (flip $\to$ agree), the mover starts from the nothink-flip side, and moves to meet the anchor at the same boundary; the two sides converge in confidence and the within-pair gap collapses. In cell $c$ (agree $\to$ flip), the mover starts at the same level as the anchor, and moves toward the opposite boundary; the two sides commit to opposite boundaries and the gap saturates. The depth-resolved confidence buildup in $|\text{CDPG}_k|$ is therefore a byproduct of thinking sharpening each side's prediction confidence: whether this sharpening creates bias (cell $c$) or resolves it (cell $b$) depends only on \emph{whether the evidence the trace accumulates is itself biased}. The primary role of thinking is to strengthen prediction confidence; the unavoidable side effect is that when the accumulated evidence carries bias, the bias is sharpened in lockstep.

\subsection{Pair-state asymmetry and group-level blindness}\label{subsec:transition}

\begin{table*}[ht]
\centering
\renewcommand{\arraystretch}{1.15}
\setlength{\tabcolsep}{4pt}
\small
\begin{tabular}{@{}l c rrr c rrr c rrr@{}}
\toprule
              & & \multicolumn{3}{c}{\textbf{Adult}} & & \multicolumn{3}{c}{\textbf{COMPAS}} & & \multicolumn{3}{c}{\textbf{Credit}} \\
\cmidrule(lr){3-5}\cmidrule(lr){7-9}\cmidrule(lr){11-13}
$A$           & & $P(Y|Y)$ & $P(Y|N)$ & $\Delta P(Y)$ & & $P(Y|Y)$ & $P(Y|N)$ & $\Delta P(Y)$ & & $P(Y|Y)$ & $P(Y|N)$ & $\Delta P(Y)$ \\
\midrule
$\alpha$           & & $0.889$  & $0.076$  & $+0.033$ & & $0.998$  & $0.281$  & $+0.174$ & & $0.908$  & $0.012$  & $-0.026$ \\
$\alpha'$          & & $0.903$  & $0.077$  & $+0.032$ & & $0.997$  & $0.272$  & $+0.182$ & & $0.912$  & $0.011$  & $-0.025$ \\
\midrule
$\boldsymbol{\Delta = T^\alpha - T^{\alpha'}}$ & & $\mathbf{-0.014}$ & $\mathbf{-0.001}$ & $\mathbf{+0.001}$ & & $\mathbf{+0.001}$ & $\mathbf{+0.009}$ & $\mathbf{-0.008}$ & & $\mathbf{-0.004}$ & $\mathbf{+0.002}$ & $\mathbf{-0.001}$ \\
\bottomrule
\end{tabular}
\caption{\textbf{Group-level transition matrix $T^A$ for Qwen3-32B, stratified by protected attribute.} $\alpha/\alpha'={}$Male/Female (Adult, Credit), White/Non-White (COMPAS); $P(Y|Y)$ and $P(Y|N)$ denote the probability of transitioning to a Y prediction from a Y or N prediction respectively. $\Delta P(Y) = P(Y)_{\text{th}} - P(Y)_{\text{nt}}$ measures the overall shift in the Yes-prediction rate (Y = earns over \$50K for Adult, will recidivate for COMPAS, will default for Credit). Other models in Appendix~\ref{app:per-instance}.}
\label{tab:transition_stratified}
\end{table*}

\begin{table*}[ht]
\centering
\renewcommand{\arraystretch}{1.15}
\setlength{\tabcolsep}{4pt}
\footnotesize
\begin{tabular}{@{}l rrr r @{\hspace{14pt}} rrr r @{\hspace{14pt}} rrr r@{}}
\toprule
              & \multicolumn{4}{c}{\textbf{Adult}}                                                  & \multicolumn{4}{c}{\textbf{COMPAS}}                                                & \multicolumn{4}{c}{\textbf{Credit}}                                                \\
\cmidrule(lr){2-5}\cmidrule(lr){6-9}\cmidrule(lr){10-13}
from          & $\mathrm{NN}$ & $\mathrm{YY}$ & $\mathrm{M}$  & $n$       & $\mathrm{NN}$ & $\mathrm{YY}$ & $\mathrm{M}$  & $n$       & $\mathrm{NN}$ & $\mathrm{YY}$ & $\mathrm{M}$  & $n$       \\
\midrule
$\mathrm{NN}$ & $0.894$ & $0.035$ & $0.071$ & $3{,}720$ & $0.686$ & $0.207$ & $0.108$ & $3{,}157$ & $0.981$ & $0.004$ & $0.015$ & $3{,}170$ \\
$\mathrm{YY}$ & $0.050$ & $0.866$ & $0.084$ & $1{,}173$ & $0.001$ & $0.996$ & $0.002$ & $1{,}698$ & $0.068$ & $0.899$ & $0.033$ & $1{,}802$ \\
$\mathrm{M}$  & $0.290$ & $0.411$ & $0.299$ & $107$     & $0.000$ & $0.986$ & $0.014$ & $145$     & $0.917$ & $0.042$ & $0.042$ & $24$      \\
\midrule
\textbf{total}& $3{,}415$ & $1{,}191$ & \cellcolor{salmon!22}$\mathbf{394}$ & $5{,}000$ & $2{,}167$ & $2{,}487$ & \cellcolor{salmon!22}$\mathbf{346}$ & $5{,}000$ & $3{,}256$ & $1{,}634$ & \cellcolor{salmon!22}$\mathbf{106}$ & $4{,}996$ \\
\bottomrule
\end{tabular}
\caption{\textbf{Pair-state transition matrix $T^{\mathrm{pair}}$ for Qwen3-32B (forward: nothink~$\to$~think).} Within-row conditional probabilities; row counts $n$ and column totals shown. Highlighted $\mathrm{M}$ column totals are the post-think cell-$c$/$d$ population sizes.}
\label{tab:transition_pair_qwen3}
\end{table*}

\begin{table}[ht]
\centering
\renewcommand{\arraystretch}{1.15}
\setlength{\tabcolsep}{3pt}
\small
\begin{tabular}{@{}l l r r r r r@{}}
\toprule
\textbf{Model} & \textbf{Dataset} & $|c|$ & $c_{\alpha\uparrow}$ & $c_{\alpha'\uparrow}$ & $\text{asym}(c)$ & $\delta_c$ (\%) \\
\midrule
\multirow{3}{*}{Qwen3}
                          & Adult  & 362 & 186 & 176 & 0.03 & $+0.20$ \\
                          & COMPAS & 344 & 170 & 174 & 0.01 & $-0.08$ \\
                          & Credit & 105 &  61 &  44 & 0.16 & $+0.34$ \\
\midrule
\multirow{3}{*}{QwQ}      & Adult  & 443 & 231 & 212 & 0.04 & $+0.38$ \\
                          & COMPAS & 351 &  82 & 269 & 0.53 & \cellcolor{salmon!22}$-3.74$ \\
                          & Credit & 595 & 232 & 363 & 0.22 & \cellcolor{salmon!22}$-2.63$ \\
\midrule
\multirow{3}{*}{R1-Distill}
                          & Adult  & 646 & 337 & 309 & 0.04 & $+0.56$ \\
                          & COMPAS & 179 &  63 & 116 & 0.30 & \cellcolor{salmon!22}$-3.71$ \\
                          & Credit & 483 & 276 & 207 & 0.14 & \cellcolor{salmon!22}$+1.38$ \\
\midrule
\textbf{Pooled} & \textbf{(9 cells)} & 3508 & 1638 & 1870 & 0.07 & $-0.56$ \\
\bottomrule
\end{tabular}
\caption{\textbf{Directional decomposition of created flips (cell $c$).} $\delta_c$ reported in \%; definitions in \S\ref{subsec:transition}. Shaded cells: the four (model, dataset) settings with $\Delta\mathcal{D}_{\text{group}}{>}0$ in Table~\ref{tab:d_group}.}
\label{tab:directional_split}
\end{table}

\paragraph{Asymmetry is not caused by group-level transition differences.}
A natural first hypothesis is that the asymmetry of $|c|>|b|$ under thinking comes from the protected attribute driving different transition rates for $\alpha$ and $\alpha'$. Table~\ref{tab:transition_stratified} rules this out by evaluating the \textbf{group-level transition matrix} (\S\ref{subsec:transition-tools}): stratifying the nothink$\to$think transition matrix by the protected attribute, $T^\alpha \approx T^{\alpha'}$ on every (dataset, transition) entry, with signed gap $|T^\alpha - T^{\alpha'}| \leq 0.014$ uniformly. 
The post-think dual effect asymmetry therefore cannot originate at the group-level transition where $T^A$ is essentially balanced across $A\in\{\alpha, \alpha'\}$. A group-rate audit (such as group fairness) that reads only $T^A$ is structurally blind to whatever drives the asymmetry. Moreover, thinking usually shifts the overall Yes-prediction rate on both sides (Table~\ref{tab:transition_stratified}). \textit{This shift could be misinterpreted by ``bias mitigated'' or ``bias amplified'' if we only study one side of the groups.} This gives another potential cause of why prior studies reach opposing conclusions.

\paragraph{Pair-state transition matrix $T^{\mathrm{pair}}$: Asymmetry is rooted in starting-state population and transition probabilities.}
As a first-order analysis, assume per-side flips are independent within a pair conditional on the nothink starting state. Writing $p := P(Y|N)$ and $q := P(N|Y) = 1-P(Y|Y)$ (Table~\ref{tab:transition_stratified}; $T^\alpha \approx T^{\alpha'}$ lets us drop the side subscript), the expected cell-$c$ and cell-$b$ counts are
\begin{align*}
\mathbb{E}|c| &= 2p(1{-}p)\,|\mathrm{NN}|_{\mathrm{nt}} + 2q(1{-}q)\,|\mathrm{YY}|_{\mathrm{nt}}, \\
\mathbb{E}|b| &= [\,p(1{-}q) + q(1{-}p)\,]\,|\mathrm{M}|_{\mathrm{nt}}.
\end{align*}
Plugging the empirical $p, q$ and nothink populations into these expressions already predicts $|c|>|b|$ on every cell (Table~\ref{tab:independence_check}, ``$|c|$ ind.''~vs.\ ``$|b|$ ind.'' columns). Empirically, independence is not strictly satisfied: the two sides of a pair share most of their input prose and tend to co-flip, so the realised $T^{\mathrm{pair}}$ (Table~\ref{tab:transition_pair_qwen3} for Qwen3-32B, Tables~\ref{tab:transition_pair_qwq}, \ref{tab:transition_pair_r1distill} for the other models) deviates from the independence formula (Appendix~\ref{app:independence-bound}). Within-pair correlation thus shifts the magnitudes but not the direction: $|c|>|b|$ holds in every cell, matching the independence prediction. The dual-effect asymmetry is therefore a structural consequence of any non-trivial transition probability acting on an imbalanced nothink population.

\paragraph{Directional decomposition of cell $c$ and audit visibility.}
Cell $c$ counts the pairs that flip under thinking but is silent on which side wins. Splitting cell $c$ into $\alpha$-favoured pairs ($c_{\alpha\uparrow}{:}\,\hat y^\alpha{=}1,\hat y^{\alpha'}{=}0$) and $\alpha'$-favoured ones ($c_{\alpha'\uparrow}$, opposite direction) under the fixed convention $\alpha/\alpha'={}$Male/Female (Adult, Credit), White/Non-White (COMPAS), adapting \citet{prabhakaran2019perturbation}, gives the signed cell-$c$ tilt $\delta_c = (c_{\alpha\uparrow}-c_{\alpha'\uparrow})/N$ (Table~\ref{tab:directional_split}). Cells $a$ and $b$ have both sides of every pair agreeing under thinking, so they contribute zero to the think-arm marginal difference between $\alpha$ and $\alpha'$; the cell-level contribution to $\Delta\mathcal{D}_{\text{group}}$ therefore originates entirely from $\delta_c$ and an analogous cell-$d$ tilt that is small because $|d|$ itself is small ($\leq 68$ pairs pooled). Empirically, the four cells with the largest $|\delta_c|$ are the four cells where $|\Delta\mathcal{D}_{\text{group}}|$ in Figure~\ref{fig:delta_paired_bars} clears the visible-signal threshold; the remaining five carry $|\delta_c|\leq 0.56\%$ and yield $|\Delta\mathcal{D}_{\text{group}}|$ at the noise floor. The five noise-floor cells are not bias-free: they are cells with large $|c|$ whose tilt happens to split symmetrically between $\alpha$ and $\alpha'$. A group-rate audit reads them as $\Delta\mathcal{D}_{\text{group}}{\approx}0$ either way, so visible group disparity is the special case of asymmetric cell-$c$ tilt.

\section{Conclusion}\label{sec:conclusion}

Thinking in reasoning models does not simply resolve counterfactual bias; it acts as a confidence amplifier that both resolves and creates counterfactual flips through one mechanism. CDPG exposes this along the depth of the thinking trace: each side of a counterfactual pair is sharpened toward whichever boundary it is already drifting to, so the same dynamic resolves a flip on some pairs and creates one on others, depending only on whether the accumulated reasoning is itself biased. BTM then locates the asymmetric dual effect at the pair-state level: per-group marginal transitions shift in near-lockstep across the two protected-attribute values, so the cross-group disagreement that drives the asymmetry is invisible to any single-group audit and only surfaces in the joint pair-state transition. This is where counterfactual bias originates under thinking, and where fairness audits of reasoning models should look.     

\section{Limitations}\label{sec:limitations}

\paragraph{Scope.}
This study evaluates counterfactual fairness on three tabular binary classification tasks with binary protected attributes (sex and race). The main study covers three 32B Qwen-lineage reasoning models; a smaller-scale replication on two non-Qwen families (Appendix~\ref{app:generalization}) supports generality, but with prompt-elicited rather than native thinking traces. Three lines of investigation are outside the present scope: a fuller group-fairness treatment beyond the demographic-parity contrast reported alongside the counterfactual audit (e.g., equalised-odds and calibration criteria; planned follow-up); extension to free-form generation tasks, multi-class attributes, and multilingual settings; and intervention-based mitigation. The paper is diagnostic: it introduces two dynamic instruments (bias propagation and the pair-state transition matrix) and characterises a previously invisible failure mode; evaluation of concrete mitigation strategies against these instruments is left to future work.

\paragraph{Methodological assumptions.}
Bias propagation uses $K{=}10$ equal-token segments of the thinking trace, a granularity that trades depth resolution against compute cost ($900{,}000$ forward passes for the full study: $3$ models $\times$ $3$ datasets $\times$ $5{,}000$ pairs $\times$ $2$ sides $\times$ $10$ depths). A sensitivity check at $K \in \{5, 10, 20\}$ (Appendix~\ref{app:k-sensitivity}) finds the cell-$c$ propagation slope preserved in sign and order of magnitude, though the check covers one model family and the linear-slope summary rather than the full trajectory shape. The within-model non-thinking baseline injects an empty closed \texttt{<think></think>} block; a small fraction of generations may still emit reasoning tokens after this prefix, and we report aggregates only over completions that respect the instruction. All inference uses 4-bit AWQ quantisation.

\paragraph{Disposition.}
The study is diagnostic, not mitigating. The two dynamic instruments and the within-pair contingency decomposition jointly provide the lens and the finding; evaluation of concrete mitigation candidates against the propagation and transition-matrix objectives is the natural follow-up and is outside the scope of this paper.

\section*{Ethics Statement}
The datasets (Adult, COMPAS, Credit) are public and extensively analysed in the fairness literature; we use them for measurement only, not training or deployment. Our finding, that thinking-token reasoning can construct high-confidence counterfactual disagreements at $5{\times}$ the per-pair magnitude of corresponding reductions, itself signals that deployment of such models in high-stakes settings should not lean on the \emph{appearance} of deliberation as fairness assurance. We do not release new model weights.

\section*{Use of AI Assistants}
AI coding assistants were used to help develop and refactor parts of the experimental codebase (data-loading utilities, inference pipelines, and plotting scripts), with all logic reviewed by the authors. AI tools were also used for limited language polishing of the manuscript (grammar, phrasing, and clarity); all scientific claims, analyses, and conclusions are the authors' own.

\bibliography{ref}

\appendix

\section{Natural-Language Profile Generation}\label{app:nl_profiles}

Each tabular record is rendered into prose by a \emph{deterministic Python template}. Because the template is a pure function of the row dict, matched (\texttt{norm}, \texttt{cf}) pairs are byte-identical except for the protected-attribute adjective and the pronouns that grammatically depend on it; no phrasing-level variance can confound the counterfactual signal.

Pronoun handling has two modes: \texttt{gendered} (he/his vs.\ she/her, flipping with \texttt{sex}; used for Adult and Credit) and \texttt{neutral} (singular \emph{they} on both sides; used for COMPAS, whose protected attribute \texttt{race} has no English pronoun system). Verbs and possessives are toggled accordingly. The templates are reproduced verbatim below.

\paragraph{Adult template.}
\begin{quote}
\begin{lstlisting}[basicstyle=\ttfamily\small]
This person is a {age}-year-old {sex}. {His} marital status is
{marital_status}. {He} work{s} in the {workclass} sector as a
{occupation} worker, {hours_per_week} hours per week. {He}
{has} a {education} education (education number {education_num}).
{He} report{s} ${capital_gain} in capital gains and
${capital_loss} in capital losses. {He} {is} from {native_country}.
\end{lstlisting}
\end{quote}

\paragraph{COMPAS template.}
\begin{quote}
\begin{lstlisting}[basicstyle=\ttfamily\small]
This defendant is a {age}-year-old {race} person. They have
{juvenile_felony_count} prior juvenile felony offense(s),
{juvenile_misdemeanor_count} prior juvenile misdemeanor offense(s),
and {juvenile_other_count} other prior juvenile offense(s). They
have {priors_count} prior adult criminal offense(s). The current
charge is a {charge_degree}.
\end{lstlisting}
\end{quote}

\paragraph{Credit template.}
\begin{quote}
\begin{lstlisting}[basicstyle=\ttfamily\small]
This person is a {age}-year-old {sex}. {His} marital status is
{marital_status} and {his} education level is {education_label}.
{He} {has} a credit limit of NT${credit_limit}. {His} statement
balances over the past 6 months were ${bill_1}, ${bill_2}, ...,
${bill_6}, and {he} paid ${paid_1}, ${paid_2}, ..., ${paid_6},
respectively. {He} {had_or_had_no} late payments in the past
6 months.
\end{lstlisting}
\end{quote}
where \texttt{education\_label} maps the UCI numeric education code via $\{1{:}\text{graduate school},\,2{:}\text{university},\,3{:}\text{high school},\,4{:}\text{other education}\}$ and unmapped codes render as \texttt{unspecified education}.

Counterfactual instances are generated by code: a single dictionary field is flipped (\texttt{sex} $\in$ \{Male, Female\} for Adult and Credit; \texttt{race} $\in$ \{White, Non-White\} for COMPAS) and the same template is re-rendered. Because the template is a pure function of the row dict, all non-protected substrings in the (\texttt{norm}, \texttt{cf}) prose are byte-identical except for the protected-attribute adjective and (for \texttt{gendered}) the pronouns that grammatically depend on it.

\section{CDPG measurement pipeline}\label{app:pipeline}

\begin{figure*}[h]
    \centering
    \resizebox{\textwidth}{!}{
\begin{tikzpicture}[
    font=\small,
    box/.style={
        draw, rounded corners=2pt, inner sep=4pt, align=center,
        minimum height=10mm, minimum width=22mm
    },
    qbox/.style={box, fill=blue!7, draw=blue!50!black},
    cfbox/.style={box, fill=red!7, draw=red!60!black},
    thinkbox/.style={
        draw, rounded corners=2pt, inner sep=3pt, align=center,
        fill=gray!8, font=\scriptsize\ttfamily
    },
    seg/.style={
        draw, rectangle, minimum width=4mm, minimum height=4.5mm,
        inner sep=0pt, font=\tiny
    },
    arrow/.style={-{Stealth[length=2.2mm]}, thick},
    label/.style={font=\scriptsize\itshape, gray!50!black},
]

\node[qbox] (qa) at (0,0) {$Q^a$\\\scriptsize age 38, sales,\\\scriptsize sex = \textbf{Male}};
\node[cfbox, below=4mm of qa] (qb) {$Q^{a'}$\\\scriptsize age 38, sales,\\\scriptsize sex = \textbf{Female}};

\node[thinkbox, right=12mm of qa] (ta) {%
\begin{tabular}{c}
{\scriptsize\textsc{thinking trace} $T^a$}\\[1pt]
\foreach \lbl/\op [count=\i] in {1/1.0,2/1.0,{\cdots}/1.0,k/1.0,{\cdots}/0.25,K/0.25}{%
  \tikz \node[seg, fill=blue!\the\numexpr 15+10*\i\relax!white, opacity=\op]{\scriptsize$\lbl$};}
\end{tabular}};
\node[thinkbox, right=12mm of qb] (tb) {%
\begin{tabular}{c}
{\scriptsize\textsc{thinking trace} $T^{a'}$}\\[1pt]
\foreach \lbl/\op [count=\i] in {1/1.0,2/1.0,{\cdots}/1.0,k/1.0,{\cdots}/0.25,K/0.25}{%
  \tikz \node[seg, fill=red!\the\numexpr 15+10*\i\relax!white, opacity=\op]{\scriptsize$\lbl$};}
\end{tabular}};

\node[box, fill=yellow!8, draw=orange!60!black, right=10mm of ta,
      minimum width=14mm] (pa) {$P^a_k$};
\node[box, fill=yellow!8, draw=orange!60!black, right=10mm of tb,
      minimum width=14mm] (pb) {$P^{a'}_k$};

\node[box, fill=violet!10, draw=violet!60!black,
      right=14mm of $(pa)!0.5!(pb)$,
      minimum width=30mm] (copg)
      {$\mathrm{CDPG}_k = P^a_k - P^{a'}_k$};

\draw[arrow] (qa) -- (ta);
\draw[arrow] (qb) -- (tb);
\draw[arrow] (ta) -- (pa);
\draw[arrow] (tb) -- (pb);
\draw[arrow] (pa.east) -- (copg.north west);
\draw[arrow] (pb.east) -- (copg.south west);

\end{tikzpicture}}
    \caption{\textbf{CDPG measurement pipeline.} Counterfactual paired inputs $(Q^\alpha, Q^{\alpha'})$ are passed through the same reasoning model; each thinking trace is split into $K{=}10$ equal-token segments. At every prefix depth $k$ we read the answer-token softmax $P^A_k{=}P(\hat y{=}1\mid Q^A, S^A_{1:k})$ and report $\mathrm{CDPG}_k{=}P^\alpha_k{-}P^{\alpha'}_k$.}\label{fig:pipeline}
\end{figure*}
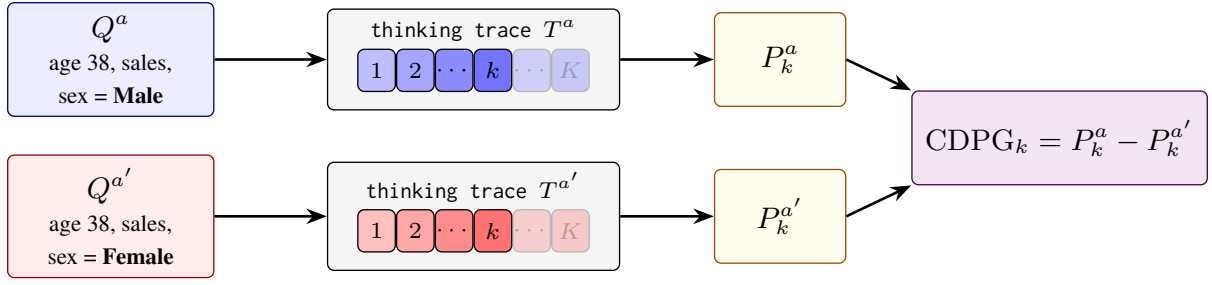

\section{Sensitivity of CDPG to the segmentation granularity $K$}\label{app:k-sensitivity}

\begin{figure}[ht]
    \centering
    \includegraphics[width=\columnwidth]{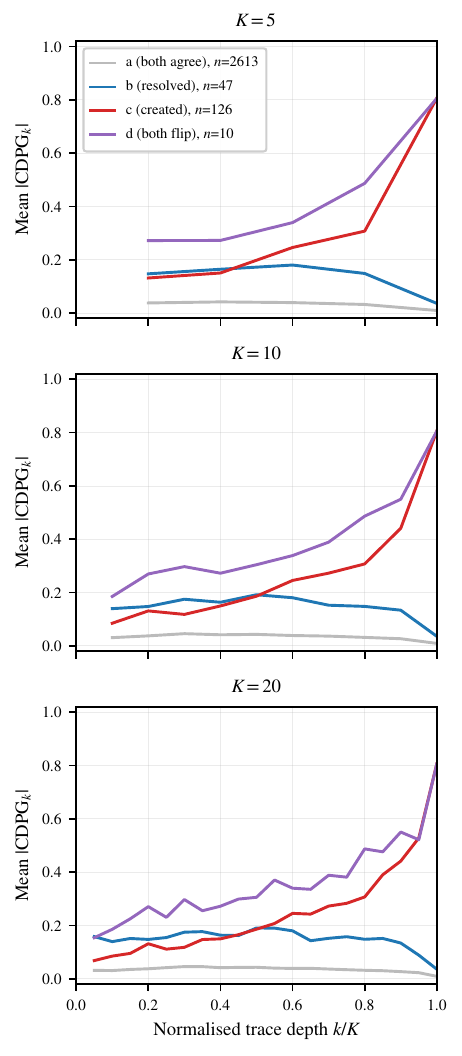}
    \caption{\textbf{CDPG trajectories are robust to the segmentation choice.} Mean $|\mathrm{CDPG}_k|$ per contingency cell as a function of normalised depth $k/K$ for $K\in\{5,10,20\}$ (Qwen3-32B, rerun subsample pooled across the three datasets; per-cell pair counts in the legend).}
    \label{fig:cdpg_k_sensitivity}
\end{figure}

The CDPG instrument partitions each thinking trace into $K{=}10$ equal-token segments (\S\ref{sec:results}). To check that the depth-wise conclusions do not depend on this choice, we recompute the trajectories at $K \in \{5, 10, 20\}$ for Qwen3-32B on all three datasets (Figure~\ref{fig:cdpg_k_sensitivity}) and fit a linear slope to the cell-$c$ mean $|\mathrm{CDPG}_k|$ trajectory against normalised depth $k/K$ (Table~\ref{tab:cdpg_k_sensitivity}).

\begin{table}[ht]
\centering
\renewcommand{\arraystretch}{1.15}
\small
\begin{tabular}{l r r r r}
\toprule
& & \multicolumn{3}{c}{\textbf{Cell-$c$ propagation slope}} \\
\cmidrule(lr){3-5}
\textbf{Dataset} & $n_c$ & $K{=}5$ & $K{=}10$ & $K{=}20$ \\
\midrule
Adult  & 54 & 0.78 & 0.65 & 0.59 \\
COMPAS & 56 & 0.77 & 0.62 & 0.54 \\
Credit & 16 & 0.61 & 0.42 & 0.32 \\
\bottomrule
\end{tabular}
\caption{\textbf{CDPG segmentation sensitivity for Qwen3-32B.} Fitted linear slope of the cell-$c$ mean $|\mathrm{CDPG}_k|$ trajectory against normalised depth $k/K$, recomputed at $K\in\{5,10,20\}$; $n_c$ is the number of cell-$c$ pairs in the rerun sample.}
\label{tab:cdpg_k_sensitivity}
\end{table}

The propagation slope for cell $c$ stays positive and of the same order of magnitude at every $K$ on all three datasets, so the depth-wise amplification that defines cell $c$ is robust to the segmentation choice. The mild decrease with $K$ is expected: $|\mathrm{CDPG}_k|$ saturates toward the end of the trace, so a linear fit over finer segments averages in more of the plateau. The $K{=}10$ conclusions reported in the main text do not depend on the specific choice of $K$.

\section{Generalization beyond the Qwen lineage}\label{app:generalization}

The three main-study models share a Qwen pretraining lineage (QwQ-32B and Qwen3-32B directly; R1-Distill via a Qwen student). To test whether the asymmetric dual effect is a lineage artefact, we replicate the audit on two model families with different pretraining lineages and sizes: gemma-2-27B (Google) and Llama-3.1-8B (Meta), at $N{=}1{,}000$ sampled records per (model, dataset) setting. Neither model exposes a native \texttt{<think>} channel, so the thinking trace is elicited by prompt and scored with the identical CDPG pipeline; the non-thinking arm and all metrics are unchanged from \S\ref{sec:results}.

\begin{table*}[ht]
\centering
\renewcommand{\arraystretch}{1.15}
\setlength{\tabcolsep}{5pt}
\small
\begin{tabular}{l l r r r r r r r r r}
\toprule
& & & \multicolumn{4}{c}{\textbf{Contingency cells (counts)}} & & \textbf{McNemar} & \multicolumn{2}{c}{$\overline{\mathcal{D}_{\text{disp}}}$ \textbf{(think$_K$)}} \\
\cmidrule(lr){4-7}\cmidrule(lr){9-9}\cmidrule(lr){10-11}
\textbf{Model (family)} & \textbf{Dataset} & $N$ & $a$ & $b$ & $c$ & $d$ & $c:b$ & $p$ & cell $c$ & cell $b$ \\
\midrule
\multirow{3}{*}{gemma-2-27B (Google)}
  & Adult  &  997 & 928 & 11 &  54 & 4 & $4.9\times$ & $6.0{\times}10^{-8}$  & \textbf{0.886} & 0.006 \\
  & COMPAS$^{\dagger}$ &  704 & 702 &  0 &   2 & 0 & ---         & $0.5$                 & 0.446 & --- \\
  & Credit & 1000 & 942 &  0 &  58 & 0 & ---         & $6.9{\times}10^{-18}$ & \textbf{0.940} & --- \\
\midrule
\multirow{3}{*}{Llama-3.1-8B (Meta)}
  & Adult  &  889 & 756 & 20 & 109 & 4 & $5.5\times$ & $5.1{\times}10^{-16}$ & \textbf{0.684} & 0.095 \\
  & COMPAS & 1000 & 842 & 18 & 140 & 0 & $7.8\times$ & $1.3{\times}10^{-24}$ & \textbf{0.293} & 0.106 \\
  & Credit &  969 & 685 & 33 & 251 & 0 & $7.6\times$ & $1.1{\times}10^{-42}$ & \textbf{0.279} & 0.084 \\
\midrule
\textbf{Pooled} & \textbf{(6 cells)} & 5559 & 4855 & 82 & 614 & 8 & $7.5\times$ & $1.3{\times}10^{-101}$ & & \\
\bottomrule
\end{tabular}
\caption{\textbf{The asymmetric dual effect replicates on two non-Qwen model families.} Same protocol and metrics as Table~\ref{tab:dual_effect_main}, at $N{=}1{,}000$ sampled records per setting ($N$ retains pairs parsing on both sides of both arms); the thinking trace is elicited by prompt since neither model exposes a native \texttt{<think>} channel. $\overline{\mathcal{D}_{\text{disp}}}$ (think$_K$) is the mean per-pair counterfactual disparity (Eq.~\ref{eq:ddisp}) at full depth within cells $c$ and $b$. $^{\dagger}$gemma-2-27B declines the COMPAS recidivism question on a large share of records (296 of $1{,}000$ pairs unparsed), leaving too few discordant pairs for a powered test; the row is shown for completeness and included in the pooled counts.}
\label{tab:generalization_models}
\end{table*}

Table~\ref{tab:generalization_models} shows the dual effect persists: created flips outnumber resolved flips in every setting with a non-trivial number of discordant pairs, at ratios ($4.9\times$--$7.8\times$, pooled $7.5\times$) comparable to the nine main settings, and McNemar's exact test rejects $b{=}c$ at $p \leq 6.0{\times}10^{-8}$ in each powered setting. The confidence signature also reproduces: created flips are committed at high per-pair disparity ($\overline{\mathcal{D}_{\text{disp}}}$ up to $0.94$ on gemma-Credit) while resolved-cell disparities collapse toward zero. The asymmetric dual effect is therefore not specific to the Qwen lineage, the 32B scale, or a native thinking channel.

\section{Inference Prompt Format}\label{app:prompt_format}

Table~\ref{tab:prompt_examples} shows the prompts used for the thinking and non-thinking conditions on the Adult dataset. The dataset-specific question line is the only piece that varies across datasets: \texttt{"Does this person earn more than \$50K per year?"} for Adult, \texttt{"Will this person re-offend within two years?"} for COMPAS, and \texttt{"Will this person default on their next credit card payment?"} for Credit.

\begin{table*}[ht]
    \centering
    \renewcommand{\arraystretch}{1.4}
    \begin{tabular}{p{2.6cm} | p{13cm}}
        \toprule
        \textbf{Condition} & \textbf{Example prompt (Adult)} \\
        \midrule
        \textbf{Thinking mode} &
        \texttt{\{description\}\newline\newline
        Question: Does this person earn more than \$50K per year?\newline
        Answer with a single word: Yes or No.}
        \\
        \midrule
        \textbf{Non-thinking mode} &
        \texttt{[Same user message as above. The chat template is rendered locally and the assistant turn is pre-filled with a closed empty \texttt{<think>\textbackslash n\textbackslash n</think>\textbackslash n\textbackslash n} block, so generation resumes after \texttt{</think>} and the model skips the reasoning phase.]} \\
        \bottomrule
    \end{tabular}
    \caption{Prompt formats for the thinking and non-thinking arms on the Adult dataset. \texttt{\{description\}} is the natural-language profile from Appendix~\ref{app:nl_profiles}.}
    \label{tab:prompt_examples}
\end{table*}

\section{Absolute disparity values per (model, dataset) cell}\label{app:disparity_values}
Figure~\ref{fig:delta_paired_bars} in the main text reports only the per-cell change $\Delta = \mathcal{D}^{\mathrm{Think}} - \mathcal{D}^{\mathrm{No\text{-}think}}$ for compactness. Table~\ref{tab:d_group} reports the corresponding absolute disparities $\mathcal{D}_{\mathrm{cf}}$ and $\mathcal{D}_{\mathrm{group}}$ under each arm.

\begin{table*}[ht]
\centering
\renewcommand{\arraystretch}{1.15}
\setlength{\tabcolsep}{5pt}
\small
\begin{tabular}{@{}l l r r@{\hspace{8pt}}r r@{\hspace{8pt}}r r@{\hspace{8pt}}r@{}}
\toprule
& & & \multicolumn{2}{c}{\textbf{Think}} & \multicolumn{2}{c}{\textbf{No-think}} & \multicolumn{2}{c}{$\boldsymbol{\Delta}$} \\
\cmidrule(lr){4-5}\cmidrule(lr){6-7}\cmidrule(lr){8-9}
\textbf{Model} & \textbf{Dataset} & $N$ & $\mathcal{D}_{\text{cf}}$ & $\mathcal{D}_{\text{group}}$ & $\mathcal{D}_{\text{cf}}$ & $\mathcal{D}_{\text{group}}$ & $\Delta\mathcal{D}_{\text{cf}}$ & $\Delta\mathcal{D}_{\text{group}}$ \\
\midrule
\multirow{3}{*}{Qwen3}
                          & Adult  & 5000 & 0.079 & 0.006 & 0.021 & 0.012 & \cellcolor{salmon!22}$+0.058$ & \cellcolor{teal!20}$-0.006$ \\
                          & COMPAS & 5000 & 0.069 & 0.001 & 0.029 & 0.010 & \cellcolor{salmon!22}$+0.040$ & \cellcolor{teal!20}$-0.009$ \\
                          & Credit & 4996 & 0.021 & 0.004 & 0.005 & 0.004 & \cellcolor{salmon!22}$+0.016$ & $\phantom{+}0.000$ \\
\midrule
\multirow{3}{*}{QwQ}
                          & Adult  & 5000 & 0.090 & 0.004 & 0.032 & 0.021 & \cellcolor{salmon!22}$+0.058$ & \cellcolor{teal!20}$-0.017$ \\
                          & COMPAS & 5000 & 0.070 & 0.037 & 0.003 & 0.003 & \cellcolor{salmon!22}$+0.067$ & \cellcolor{salmon!22}$+0.034$ \\
                          & Credit & 4984 & 0.120 & 0.026 & 0.025 & 0.021 & \cellcolor{salmon!22}$+0.095$ & \cellcolor{salmon!22}$+0.005$ \\
\midrule
\multirow{3}{*}{R1-Distill}
                          & Adult  & 4999 & 0.133 & 0.006 & 0.020 & 0.008 & \cellcolor{salmon!22}$+0.113$ & \cellcolor{teal!20}$-0.002$ \\
                          & COMPAS & 1430 & 0.125 & 0.037 & 0.003 & 0.003 & \cellcolor{salmon!22}$+0.122$ & \cellcolor{salmon!22}$+0.034$ \\
                          & Credit & 4995 & 0.097 & 0.014 & 0.002 & 0.000 & \cellcolor{salmon!22}$+0.095$ & \cellcolor{salmon!22}$+0.014$ \\
\bottomrule
\end{tabular}
\caption{\textbf{Counterfactual disparity $\mathcal{D}_{\text{cf}}$ (Eq.~\ref{eq:dcf}) and demographic-parity gap $\mathcal{D}_{\text{group}}$ (Eq.~\ref{eq:dgroup}) under the Think and No-think arms, with within-cell change $\Delta = \mathcal{D}^{\text{Think}} - \mathcal{D}^{\text{No-think}}$.} Positive $\Delta$ amplifies the bias, negative $\Delta$ mitigates (palette from Table~\ref{tab:cell_definitions}). $N$ matches Table~\ref{tab:dual_effect_main}.}
\label{tab:d_group}
\end{table*}

\section{Group-level transition matrix: detailed breakdown}\label{app:per-instance}

Table~\ref{tab:transition_stratified} in the main text reports the group-level transition matrix $T^A_{ij} = P(\hat y_{\text{think}}{=}j \mid \hat y_{\text{nothink}}{=}i,\, A)$ for $A\in\{\alpha,\alpha'\}$ within each dataset for Qwen3-32B. Tables~\ref{tab:transition_stratified_qwq} and~\ref{tab:transition_stratified_r1distill} extend the same stratification to QwQ-32B and DeepSeek-R1-Distill-Qwen-32B, including the per-group marginal drift $\Delta P(Y) = P(Y)_{\text{th}} - P(Y)_{\text{nt}}$. Across all three models the per-(model, dataset, direction) gap between $T^\alpha$ and $T^{\alpha'}$ remains tightly bounded: $|\Delta|\leq 0.014$ on Qwen3-32B, $|\Delta|\leq 0.017$ on QwQ-32B (largest on Credit $P(Y|Y)$), and $|\Delta|\leq 0.012$ on R1-Distill-Qwen-32B. These gaps are an order of magnitude smaller than the cell-$c$ pair-state tilts documented in the main text (e.g., QwQ-COMPAS $\mathrm{asym}{=}0.53$); the pair-level cell-$c$ tilts therefore cannot be inherited from group-level bias, since the joint property captured by $T^{\mathrm{pair}}$ (Appendix~\ref{app:transition}) is what places the constraint on whether the two paired instances of a (norm, cf) pair transition consistently with each other.

\begin{table*}[ht]
\centering
\renewcommand{\arraystretch}{1.15}
\setlength{\tabcolsep}{4pt}
\small
\begin{tabular}{@{}l c rrr c rrr c rrr@{}}
\toprule
              & & \multicolumn{3}{c}{\textbf{Adult}} & & \multicolumn{3}{c}{\textbf{COMPAS}} & & \multicolumn{3}{c}{\textbf{Credit}} \\
\cmidrule(lr){3-5}\cmidrule(lr){7-9}\cmidrule(lr){11-13}
$A$           & & $P(Y|Y)$ & $P(Y|N)$ & $\Delta P(Y)$ & & $P(Y|Y)$ & $P(Y|N)$ & $\Delta P(Y)$ & & $P(Y|Y)$ & $P(Y|N)$ & $\Delta P(Y)$ \\
\midrule
$\alpha$           & & $0.980$ & $0.300$ & $+0.249$ & & $1.000$ & $0.525$ & $+0.525$ & & $0.977$ & $0.387$ & $+0.361$ \\
$\alpha'$          & & $0.988$ & $0.302$ & $+0.255$ & & $1.000$ & $0.515$ & $+0.513$ & & $0.961$ & $0.381$ & $+0.355$ \\
\midrule
$\boldsymbol{\Delta = T^\alpha - T^{\alpha'}}$ & & $\mathbf{-0.008}$ & $\mathbf{-0.002}$ & $\mathbf{-0.006}$ & & $\mathbf{+0.000}$ & $\mathbf{+0.010}$ & $\mathbf{+0.011}$ & & $\mathbf{+0.017}$ & $\mathbf{+0.006}$ & $\mathbf{+0.006}$ \\
\bottomrule
\end{tabular}
\caption{\textbf{Group-level transition matrix $T^A$ for QwQ-32B, stratified by protected attribute.} Same conventions as Table~\ref{tab:transition_stratified}.}
\label{tab:transition_stratified_qwq}
\end{table*}

\begin{table*}[ht]
\centering
\renewcommand{\arraystretch}{1.15}
\setlength{\tabcolsep}{4pt}
\small
\begin{tabular}{@{}l c rrr c rrr c rrr@{}}
\toprule
              & & \multicolumn{3}{c}{\textbf{Adult}} & & \multicolumn{3}{c}{\textbf{COMPAS}} & & \multicolumn{3}{c}{\textbf{Credit}} \\
\cmidrule(lr){3-5}\cmidrule(lr){7-9}\cmidrule(lr){11-13}
$A$           & & $P(Y|Y)$ & $P(Y|N)$ & $\Delta P(Y)$ & & $P(Y|Y)$ & $P(Y|N)$ & $\Delta P(Y)$ & & $P(Y|Y)$ & $P(Y|N)$ & $\Delta P(Y)$ \\
\midrule
$\alpha$           & & $0.968$ & $0.196$ & $+0.138$ & & $1.000$ & $0.215$ & $+0.203$ & & $1.000$ & $0.382$ & $+0.381$ \\
$\alpha'$          & & $0.977$ & $0.201$ & $+0.147$ & & $1.000$ & $0.203$ & $+0.193$ & & $1.000$ & $0.392$ & $+0.391$ \\
\midrule
$\boldsymbol{\Delta = T^\alpha - T^{\alpha'}}$ & & $\mathbf{-0.009}$ & $\mathbf{-0.005}$ & $\mathbf{-0.009}$ & & $\mathbf{+0.000}$ & $\mathbf{+0.012}$ & $\mathbf{+0.010}$ & & $\mathbf{+0.000}$ & $\mathbf{-0.010}$ & $\mathbf{-0.010}$ \\
\bottomrule
\end{tabular}
\caption{\textbf{Group-level transition matrix $T^A$ for DeepSeek-R1-Distill-Qwen-32B, stratified by protected attribute.} Same conventions as Table~\ref{tab:transition_stratified}.}
\label{tab:transition_stratified_r1distill}
\end{table*}

Pooled across $84{,}428$ per-instance observations under a per-side parseable filter (an instance is retained iff its own think and nothink arms both emit Yes/No; this is looser than the pair-level filter used in the main text, since one side of a pair can be unparseable without dropping the other), the marginal $T = \mathbb{E}_A[T^A]$ is
\begin{center}\small
\begin{tabular}{l rrr}
\toprule
 & think${=}Y$ & think${=}N$ & total \\
\midrule
nothink${=}Y$ & $13{,}891$ ($0.952$) & $705$ ($0.048$) & $14{,}596$ \\
nothink${=}N$ & $20{,}216$ ($\mathbf{0.289}$) & $49{,}616$ ($0.711$) & $69{,}832$ \\
\bottomrule
\end{tabular}
\end{center}
Thinking exerts a strong group-level $N{\to}Y$ push: the $N{\to}Y$ count outnumbers the $Y{\to}N$ count by $20{,}216/705 = 28.7{\times}$, and the two off-diagonals of $T$ are themselves asymmetric ($0.048$ vs.\ $0.289$, a $6{\times}$ ratio). Thinking moves predictions toward $Y$ for both protected attribute values, but operates more aggressively on $N$ baselines than on $Y$ baselines.

\section{Independence-bound check per (model, dataset) cell}\label{app:independence-bound}

Table~\ref{tab:independence_check} compares the independence-based predictions for $|c|$ and $|b|$ (\S\ref{subsec:transition}) against the empirical counts in each of the nine (model, dataset) cells. For each cell, $p = T^A_{NY}$ and $q = T^A_{YN}$ are read from the group-level transition matrices (Tables~\ref{tab:transition_stratified}, \ref{tab:transition_stratified_qwq}, \ref{tab:transition_stratified_r1distill}, averaged across $\alpha$ and $\alpha'$ since $T^\alpha \approx T^{\alpha'}$), and the nothink-state populations $|\mathrm{NN}|_{\mathrm{nt}}, |\mathrm{YY}|_{\mathrm{nt}}, |\mathrm{M}|_{\mathrm{nt}}$ are read from the row totals of the pair-state matrices (Tables~\ref{tab:transition_pair_qwen3}, \ref{tab:transition_pair_qwq}, \ref{tab:transition_pair_r1distill}).

\begin{table*}[ht]
\centering
\renewcommand{\arraystretch}{1.15}
\setlength{\tabcolsep}{5pt}
\small
\begin{tabular}{@{}ll rrr c rrr c rr@{}}
\toprule
              &        & \multicolumn{3}{c}{\textbf{$|c|$ (created)}} & & \multicolumn{3}{c}{\textbf{$|b|$ (resolved)}} & & \multicolumn{2}{c}{\textbf{$|c|/|b|$}} \\
\cmidrule(lr){3-5}\cmidrule(lr){7-9}\cmidrule(lr){11-12}
Model         & Dataset & ind. & actual & ratio & & ind. & actual & ratio & & ind. & actual \\
\midrule
Qwen3-32B     & Adult   & $744$  & $363$ & $2.05\times$ & & $18$  & $75$  & $0.24\times$ & & $41$  & $4.8$  \\
Qwen3-32B     & COMPAS  & $1272$ & $344$ & $3.69\times$ & & $40$  & $143$ & $0.28\times$ & & $32$  & $2.4$  \\
Qwen3-32B     & Credit  & $367$  & $107$ & $3.43\times$ & & $2$   & $23$  & $0.10\times$ & & $153$ & $4.7$  \\
\midrule
QwQ-32B       & Adult   & $1768$ & $444$ & $3.98\times$ & & $50$  & $152$ & $0.33\times$ & & $36$  & $2.9$  \\
QwQ-32B       & COMPAS  & $2487$ & $349$ & $7.13\times$ & & $8$   & $15$  & $0.52\times$ & & $319$ & $23.3$ \\
QwQ-32B       & Credit  & $2198$ & $598$ & $3.68\times$ & & $49$  & $121$ & $0.40\times$ & & $45$  & $4.9$  \\
\midrule
R1-Distill    & Adult   & $1246$ & $649$ & $1.92\times$ & & $21$  & $77$  & $0.27\times$ & & $59$  & $8.4$  \\
R1-Distill    & COMPAS  & $448$  & $179$ & $2.50\times$ & & $1$   & $4$   & $0.21\times$ & & $560$ & $44.8$ \\
R1-Distill    & Credit  & $2359$ & $487$ & $4.84\times$ & & $4$   & $10$  & $0.39\times$ & & $609$ & $48.7$ \\
\bottomrule
\end{tabular}
\caption{\textbf{Independence-bound check per (model, dataset) cell.} Independence predictions use $\mathbb{E}|c| = 2p(1{-}p)\,|\mathrm{NN}|_{\mathrm{nt}} + 2q(1{-}q)\,|\mathrm{YY}|_{\mathrm{nt}}$ and $\mathbb{E}|b| = [p(1{-}q){+}q(1{-}p)]\,|\mathrm{M}|_{\mathrm{nt}}$ with $p = P(Y|N)$ and $q = P(N|Y)$ read from Tables~\ref{tab:transition_stratified} (Qwen3-32B), \ref{tab:transition_stratified_qwq} (QwQ-32B), \ref{tab:transition_stratified_r1distill} (R1-Distill). Actual counts are read from Tables~\ref{tab:transition_pair_qwen3}, \ref{tab:transition_pair_qwq}, \ref{tab:transition_pair_r1distill}. Across all nine cells, independence \emph{overpredicts} $|c|$ by 2--7$\times$ and \emph{underpredicts} $|b|$ by 2--10$\times$, but the directional inequality $|c|>|b|$ holds under both independence and the data. The empirical ratios $|c|/|b|$ are uniformly smaller than the independence prediction (e.g., 4.8 vs.\ 41 on Qwen3-Adult): within-pair flip correlation dampens the base-count asymmetry but cannot reverse it.}
\label{tab:independence_check}
\end{table*}

Two consistent patterns appear. First, the directional inequality $|c|>|b|$ holds under both independence and the data on every cell: even purely independent per-side flips would produce the dual-effect asymmetry, given the observed nothink-state population skew. Second, the independence prediction is uniformly looser than the data: it overpredicts $|c|$ (2--7$\times$) and underpredicts $|b|$ (2--10$\times$), so the empirical $|c|/|b|$ ratios are $5$--$100\times$ smaller than the independence-bound ratios. This gap quantifies within-pair flip correlation: two instances of a counterfactual pair share most of their input prose and therefore tend to co-flip rather than flip independently. This correlation acts in the direction of dampening the base-count asymmetry, but cannot reverse it.

\section{Per-model pair-state transition matrices}\label{app:transition}

The pair-state transition matrices $T^{\mathrm{pair}}$ for QwQ-32B and DeepSeek-R1-Distill-Qwen-32B are presented here, along with the per-side stratifications. The Qwen3-32B counterpart is given in the main text (Table~\ref{tab:transition_pair_qwen3}). The qualitative pattern reproduces in every (model, dataset) cell: $\mathrm{NN}$ dominates the starting distribution, $\mathrm{NN}\!\to\!\mathrm{M}$ is the dominant cell-$c$ channel, and the $\mathrm{NN}\!\to\!\mathrm{YY}$ en-bloc rate varies by model.

\paragraph{Model heterogeneity is in the channel mix.}
All three models exhibit the same $|c|\!\gg\!|b|$ mechanism (\S\ref{subsec:transition}); what varies is the relative weight of two channels out of $\mathrm{NN}$: en-bloc push to $\mathrm{YY}$ versus split into $\mathrm{M}$. Qwen3-32B is the most $\mathrm{NN}$-inert, with $\mathrm{NN}$-row stickiness $89\%/69\%/98\%$ across Adult/COMPAS/Credit (Table~\ref{tab:transition_pair_qwen3}) and the en-bloc $\mathrm{NN}\!\to\!\mathrm{YY}$ channel never exceeding $21\%$. QwQ-32B is en-bloc-aggressive, with $\mathrm{NN}\!\to\!\mathrm{YY}$ at $24\%/48\%/31\%$ across the three datasets (Table~\ref{tab:transition_pair_qwq}). DeepSeek-R1-Distill is split-aggressive, with $\mathrm{NN}\!\to\!\mathrm{M}$ peaking at $16.4\%$ on Adult (Table~\ref{tab:transition_pair_r1distill}).

\begin{table*}[h]
\centering
\renewcommand{\arraystretch}{1.15}
\setlength{\tabcolsep}{3pt}
\footnotesize
\begin{tabular}{@{}l rrrr r @{\hspace{10pt}} rrrr r @{\hspace{10pt}} rrrr r@{}}
\toprule
              & \multicolumn{5}{c}{\textbf{Adult}}                                                                          & \multicolumn{5}{c}{\textbf{COMPAS}}                                                                         & \multicolumn{5}{c}{\textbf{Credit}}                                                                         \\
\cmidrule(lr){2-6}\cmidrule(lr){7-11}\cmidrule(lr){12-16}
from          & $\mathrm{NN}$ & $\mathrm{YY}$ & $\mathrm{YN}$        & $\mathrm{NY}$        & $n$       & $\mathrm{NN}$ & $\mathrm{YY}$ & $\mathrm{YN}$        & $\mathrm{NY}$        & $n$       & $\mathrm{NN}$ & $\mathrm{YY}$ & $\mathrm{YN}$        & $\mathrm{NY}$        & $n$       \\
\midrule
$\mathrm{NN}$ & $0.659$ & $0.236$ & $\mathbf{0.055}$ & $\mathbf{0.050}$ & $4{,}149$ & $0.445$ & $0.484$ & $\mathbf{0.016}$ & $\mathbf{0.054}$ & $4{,}981$ & $0.561$ & $0.312$ & $\mathbf{0.050}$ & $\mathbf{0.077}$ & $4{,}616$ \\
$\mathrm{YY}$ & $0.004$ & $0.984$ & $\mathbf{0.006}$ & $\mathbf{0.006}$ & $690$     & $0.000$ & $1.000$ & $\mathbf{0.000}$ & $\mathbf{0.000}$ & $4$       & $0.008$ & $0.943$ & $\mathbf{0.012}$ & $\mathbf{0.037}$ & $244$     \\
$\mathrm{YN}$ & $0.045$ & $0.888$ & $0.030$          & $0.037$          & $134$     & $0.000$ & $1.000$ & $0.000$          & $0.000$          & $14$      & $0.009$ & $0.965$ & $0.009$          & $0.018$          & $114$     \\
$\mathrm{NY}$ & $0.000$ & $1.000$ & $0.000$          & $0.000$          & $27$      & $0.000$ & $1.000$ & $0.000$          & $0.000$          & $1$       & $0.000$ & $1.000$ & $0.000$          & $0.000$          & $10$      \\
\midrule
\textbf{total}& $2{,}743$ & $1{,}805$ & $235$ & $217$ & $5{,}000$ & $2{,}218$ & $2{,}431$ & $82$ & $269$ & $5{,}000$ & $2{,}594$ & $1{,}792$ & $233$ & $365$ & $4{,}984$ \\
\bottomrule
\end{tabular}
\caption{\textbf{Pair-state transition matrix $T^{\mathrm{pair}}$ for QwQ-32B, per dataset (forward: nothink~$\to$~think), with $\mathrm{M}$ split into $\mathrm{YN}$ and $\mathrm{NY}$.} Compared with Qwen3-32B (Table~\ref{tab:transition_pair_qwen3}), QwQ runs hotter on every channel: the en-bloc $\mathrm{NN}\!\to\!\mathrm{YY}$ push peaks at $48.4\%$ on COMPAS (half the originally-agree-No pairs are pushed wholesale to agree-Yes), and within $\mathrm{M}$ the cell-$c$ split tilts strongly toward Non-White on COMPAS ($\mathrm{NN}\!\to\!\mathrm{NY}=0.054$ vs $\mathrm{NN}\!\to\!\mathrm{YN}=0.016$) and toward Female on Credit ($\mathrm{NN}\!\to\!\mathrm{NY}=0.077$ vs $\mathrm{NN}\!\to\!\mathrm{YN}=0.050$); the YY side reinforces the Credit tilt ($\mathrm{YY}\!\to\!\mathrm{NY}=0.037$ vs $\mathrm{YY}\!\to\!\mathrm{YN}=0.012$).}
\label{tab:transition_pair_qwq}
\end{table*}

\begin{table*}[h]
\centering
\renewcommand{\arraystretch}{1.15}
\setlength{\tabcolsep}{3pt}
\footnotesize
\begin{tabular}{@{}l rrrr r @{\hspace{10pt}} rrrr r @{\hspace{10pt}} rrrr r@{}}
\toprule
              & \multicolumn{5}{c}{\textbf{Adult}}                                                                          & \multicolumn{5}{c}{\textbf{COMPAS}}                                                                         & \multicolumn{5}{c}{\textbf{Credit}}                                                                         \\
\cmidrule(lr){2-6}\cmidrule(lr){7-11}\cmidrule(lr){12-16}
from          & $\mathrm{NN}$ & $\mathrm{YY}$ & $\mathrm{YN}$        & $\mathrm{NY}$        & $n$       & $\mathrm{NN}$ & $\mathrm{YY}$ & $\mathrm{YN}$        & $\mathrm{NY}$        & $n$       & $\mathrm{NN}$ & $\mathrm{YY}$ & $\mathrm{YN}$        & $\mathrm{NY}$        & $n$       \\
\midrule
$\mathrm{NN}$ & $0.728$ & $0.108$ & $\mathbf{0.087}$ & $\mathbf{0.078}$ & $3{,}715$ & $0.726$ & $0.142$ & $\mathbf{0.046}$ & $\mathbf{0.086}$ & $1{,}356$ & $0.565$ & $0.338$ & $\mathbf{0.056}$ & $\mathbf{0.042}$ & $4{,}971$ \\
$\mathrm{YY}$ & $0.007$ & $0.964$ & $\mathbf{0.012}$ & $\mathbf{0.018}$ & $1{,}186$ & $0.000$ & $1.000$ & $\mathbf{0.000}$ & $\mathbf{0.000}$ & $70$      & $0.000$ & $1.000$ & $\mathbf{0.000}$ & $\mathbf{0.000}$ & $14$      \\
$\mathrm{YN}$ & $0.059$ & $0.735$ & $0.103$          & $0.103$          & $68$      & --      & --      & --               & --               & $0$       & $0.000$ & $1.000$ & $0.000$          & $0.000$          & $5$       \\
$\mathrm{NY}$ & $0.100$ & $0.667$ & $0.133$          & $0.100$          & $30$      & $0.000$ & $1.000$ & $0.000$          & $0.000$          & $4$       & $0.000$ & $1.000$ & $0.000$          & $0.000$          & $5$       \\
\midrule
\textbf{total}& $2{,}718$ & $1{,}614$ & $348$ & $319$ & $4{,}999$ & $985$ & $266$ & $63$ & $116$ & $1{,}430$ & $2{,}808$ & $1{,}704$ & $276$ & $207$ & $4{,}995$ \\
\bottomrule
\end{tabular}
\caption{\textbf{Pair-state transition matrix $T^{\mathrm{pair}}$ for DeepSeek-R1-Distill-Qwen-32B, per dataset (forward: nothink~$\to$~think), with $\mathrm{M}$ split into $\mathrm{YN}$ and $\mathrm{NY}$.} R1-Distill has the highest aggregated $\mathrm{NN}\!\to\!\mathrm{M}$ rates of the three models (Adult: $\mathrm{NN}\!\to\!\mathrm{YN}+\mathrm{NN}\!\to\!\mathrm{NY}=0.165$, the per-model peak); the SFT-distillation regime is the most aggressive at producing pair-level splits rather than en-bloc pushes. The COMPAS row reflects $\sim 72\%$ refusal under R1-Distill's prompt sensitivity (only $1{,}430$ parseable pairs); on the parseable subset $\mathrm{NN}$ tilts strongly toward Non-White ($\mathrm{NN}\!\to\!\mathrm{NY}=0.086$ vs $\mathrm{NN}\!\to\!\mathrm{YN}=0.046$), matching QwQ on COMPAS. On Credit, the direction reverses ($\mathrm{NN}\!\to\!\mathrm{YN}=0.056$ vs $\mathrm{NN}\!\to\!\mathrm{NY}=0.042$, Male-leaning), opposite to QwQ.}
\label{tab:transition_pair_r1distill}
\end{table*}

\end{document}